Future of Humanity Institute

University of Oxford

Centre for the Study of Existential Risk

University of Cambridge

Center for a New American Security

Electronic Frontier Foundation

OpenAI

# The Malicious Use of Artificial Intelligence: Forecasting, Prevention, and Mitigation

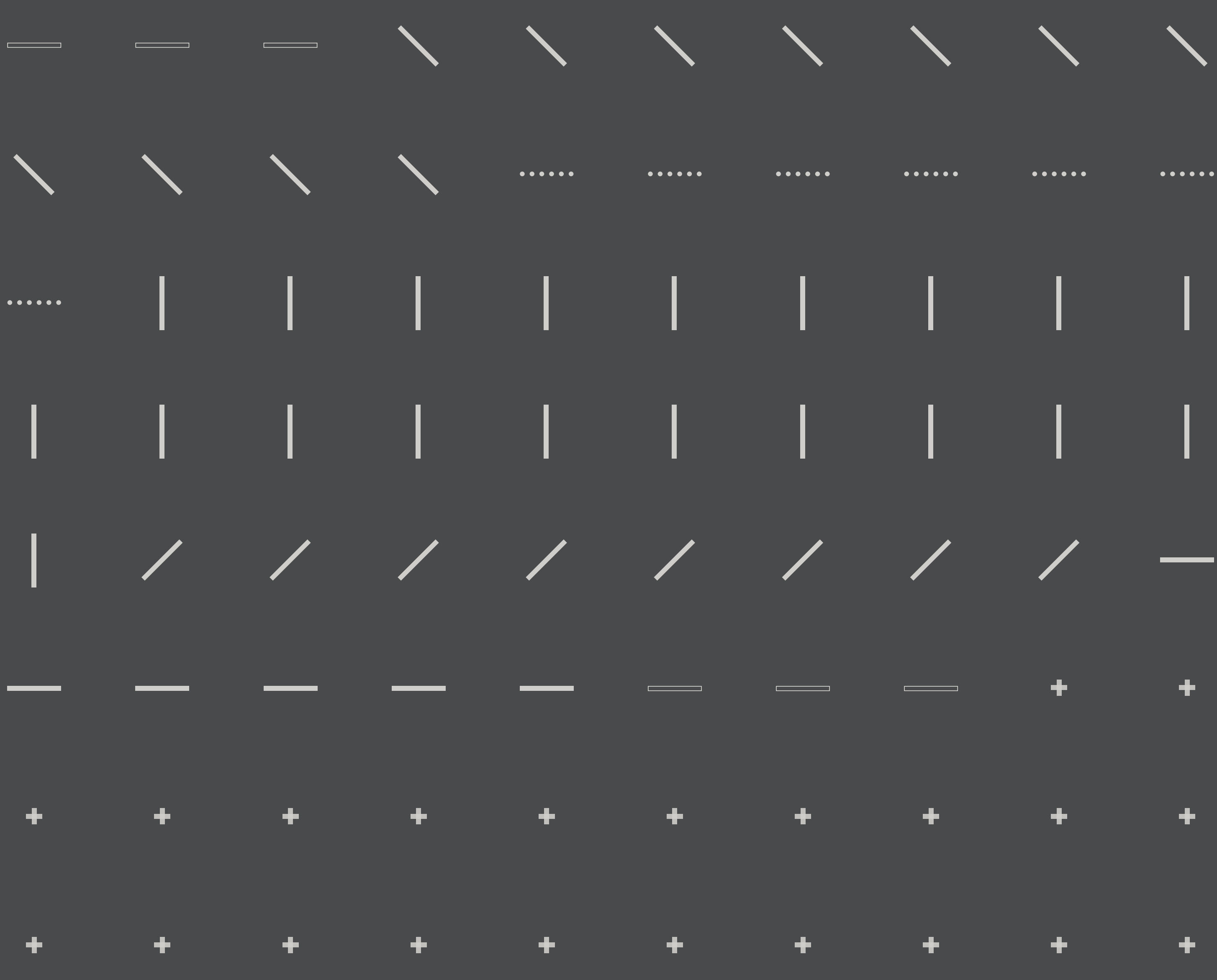

Miles Brundage[1] Shahar Avin[2] Jack Clark[3] Helen Toner[4] Peter Eckersley[5] Ben Garfinkel[6] Allan Dafoe[7] Paul Scharre[8] Thomas Zeitzoff[9] Bobby Filar[10] Hyrum Anderson[11] Heather Roff[12] Gregory C. Allen[13] Jacob Steinhardt[14] Carrick Flynn[15] Seán Ó hÉigeartaigh[16] Simon Beard[17] Haydn Belfield[18] Sebastian Farquhar[19] Clare Lyle[20] Rebecca Crootof[21] Owain Evans[22] Michael Page[23] Joanna Bryson [24] Roman Yampolskiy[25] Dario Amodei[26]

1 Corresponding author
miles.brundage@philosophy.ox.ac.uk
Future of Humanity Institute, University of Oxford; Arizona State University

2 Corresponding author,
sa478@cam.ac.uk
Centre for the Study of Existential Risk, University of Cambridge

3 OpenAI

4 Open Philanthropy Project

5 Electronic Frontier Foundation

6 Future of Humanity Institute, University of Oxford

7 Future of Humanity Institute, University of Oxford; Yale University

8 Center for a New American Security

9 American University

10 Endgame

11 Endgame

12 University of Oxford/Arizona State University/New America Foundation

13 Center for a New American Security

14 Stanford University

15 Future of Humanity Institute, University of Oxford

16 Centre for the Study of Existential Risk and Centre for the Future of Intelligence, University of Cambridge

17 Centre for the Study of Existential Risk, University of Cambridge

18 Centre for the Study of Existential Risk, University of Cambridge

19 Future of Humanity Institute, University of Oxford

20 Future of Humanity Institute, University of Oxford

21 Information Society Project, Yale University

22 Future of Humanity Institute, University of Oxford

23 OpenAI

24 University of Bath

25 University of Louisville

26 OpenAI

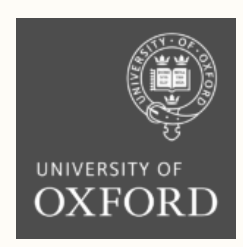

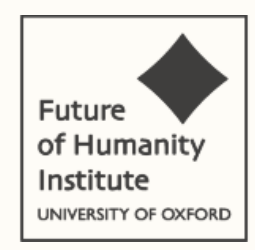

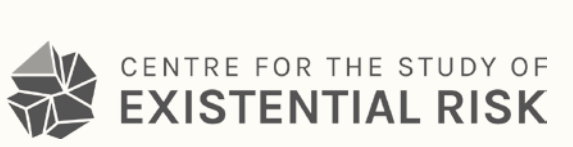

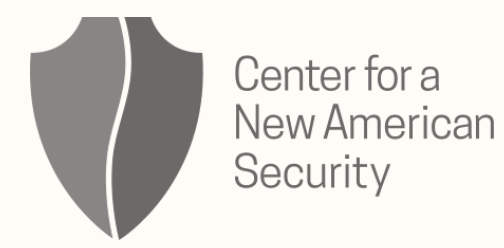

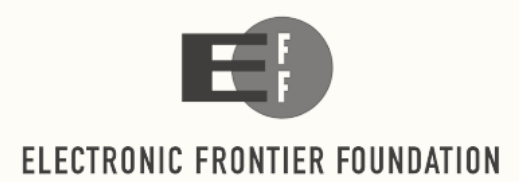

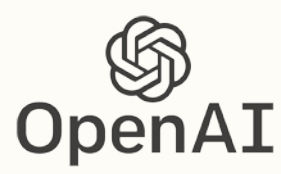

# 00 Executive Summary

Artificial intelligence and machine learning capabilities are growing at an unprecedented rate. These technologies have many widely beneficial applications, ranging from machine translation to medical image analysis. Countless more such applications are being developed and can be expected over the long term. Less attention has historically been paid to the ways in which artificial intelligence can be used maliciously. This report surveys the landscape of potential security threats from malicious uses of artificial intelligence technologies, and proposes ways to better forecast, prevent, and mitigate these threats. We analyze, but do not conclusively resolve, the question of what the long-term equilibrium between attackers and defenders will be. We focus instead on what sorts of attacks we are likely to see soon if adequate defenses are not developed.

In response to the changing threat landscape we make four high-level recommendations:

1. Policymakers should collaborate closely with technical researchers to investigate, prevent, and mitigate potential malicious uses of AI.

2. Researchers and engineers in artificial intelligence should take the dual-use nature of their work seriously, allowing misuse-related considerations to influence research priorities and norms, and proactively reaching out to relevant actors when harmful applications are foreseeable.

3. Best practices should be identified in research areas with more mature methods for addressing dual-use concerns, such as computer security, and imported where applicable to the case of AI.

4. Actively seek to expand the range of stakeholders and domain experts involved in discussions of these challenges.

As AI capabilities become more powerful and widespread, we expect the growing use of AI systems to lead to the following changes in the landscape of threats:

- Expansion of existing threats. The costs of attacks may be lowered by the scalable use of AI systems to complete tasks that would ordinarily require human labor, intelligence and expertise. A natural effect would be to expand the set of actors who can carry out particular attacks, the rate at which they can carry out these attacks, and the set of potential targets.

- Introduction of new threats. New attacks may arise through the use of AI systems to complete tasks that would be otherwise impractical for humans. In addition, malicious actors may exploit the vulnerabilities of AI systems deployed by defenders.

- Change to the typical character of threats. We believe there is reason to expect attacks enabled by the growing use of AI to be especially effective, finely targeted, difficult to attribute, and likely to exploit vulnerabilities in AI systems.

We structure our analysis by separately considering three security domains, and illustrate possible changes to threats within these domains through representative examples:

- Digital security. The use of AI to automate tasks involved in carrying out cyberattacks will alleviate the existing tradeoff between the scale and efficacy of attacks. This may expand the threat associated with labor-intensive cyberattacks (such as spear phishing). We also expect novel attacks that exploit human vulnerabilities (e.g. through the use of speech synthesis for impersonation), existing software vulnerabilities (e.g. through automated hacking), or the vulnerabilities of AI systems (e.g. through adversarial examples and data poisoning).

- Physical security. The use of AI to automate tasks involved in carrying out attacks with drones and other physical systems (e.g. through the deployment of autonomous weapons systems) may expand the threats associated with these attacks. We also expect novel attacks that subvert cyber-physical systems (e.g. causing autonomous vehicles to crash) or involve physical systems that it would be infeasible to direct remotely (e.g. a swarm of thousands of micro-drones).

- Political security. The use of AI to automate tasks involved in surveillance (e.g. analysing mass-collected data), persuasion (e.g. creating targeted propaganda), and deception (e.g. manipulating videos) may expand threats associated with privacy invasion and social manipulation. We also expect novel attacks that take advantage of an improved capacity to analyse human behaviors, moods, and beliefs on the basis of available data. These concerns are most significant in the context of authoritarian states, but may also undermine the ability of democracies to sustain truthful public debates.

In addition to the high-level recommendations listed above, we also propose the exploration of several open questions and potential interventions within four priority research areas:

- Learning from and with the cybersecurity community. At the intersection of cybersecurity and AI attacks, we highlight the need to explore and potentially implement red teaming, formal verification, responsible disclosure of AI vulnerabilities, security tools, and secure hardware.

- Exploring different openness models. As the dual-use nature of AI and ML becomes apparent, we highlight the need to reimagine norms and institutions around the openness of research, starting with pre-publication risk assessment in technical areas of special concern, central access licensing models, sharing regimes that favor safety and security, and other lessons from other dual-use technologies.

- Promoting a culture of responsibility. AI researchers and the organisations that employ them are in a unique position to shape the security landscape of the AI-enabled world. We highlight the importance of education, ethical statements and standards, framings, norms, and expectations.

- Developing technological and policy solutions. In addition to the above, we survey a range of promising technologies, as well as policy interventions, that could help build a safer future with AI. High-level areas for further research include privacy protection, coordinated use of AI for public-good security, monitoring of AI-relevant resources, and other legislative and regulatory responses.

The proposed interventions require attention and action not just from AI researchers and companies but also from legislators, civil servants, regulators, security researchers and educators. The challenge is daunting and the stakes are high.

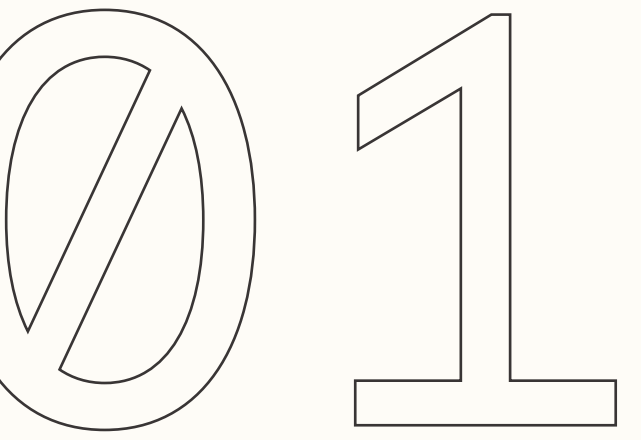

# Introduction

Artificial intelligence (AI) and machine learning (ML)[1] have progressed rapidly in recent years, and their development has enabled a wide range of beneficial applications. For example, AI is a critical component of widely used technologies such as automatic speech recognition, machine translation, spam filters, and search engines. Additional promising technologies currently being researched or undergoing small-scale pilots include driverless cars, digital assistants for nurses and doctors, and AI-enabled drones for expediting disaster relief operations. Even further in the future, advanced AI holds out the promise of reducing the need for unwanted labor, greatly expediting scientific research, and improving the quality of governance. We are excited about many of these developments, though we also urge attention to the ways in which AI can be used maliciously[2]. We analyze such risks in detail so that they can be prevented or mitigated, not just for the value of

1 AI refers to the use of digital technology to create systems that are capable of performing tasks commonly thought to require intelligence. Machine learning is variously characterized as either a sub-field of AI or a separate field, and refers to the development of digital systems that improve their performance on a given task over time through experience.

2 We define "malicious use" loosely, to include all practices that are intended to compromise the security of individuals, groups, or a society. Note that one could read much of our document under various possible perspectives on what constitutes malicious use, as the interventions and structural issues we discuss are fairly general.

preventing the associated harms, but also to prevent delays in the realization of the beneficial applications of AI.

Artificial intelligence (AI) and machine learning (ML) are altering the landscape of security risks for citizens, organizations, and states. Malicious use of AI could threaten digital security (e.g. through criminals training machines to hack or socially engineer victims at human or superhuman levels of performance), physical security (e.g. non-state actors weaponizing consumer drones), and political security (e.g. through privacy-eliminating surveillance, profiling, and repression, or through automated and targeted disinformation campaigns).

The malicious use of AI will impact how we construct and manage our digital infrastructure as well as how we design and distribute AI systems, and will likely require policy and other institutional responses. The question this report hopes to answer is: how can we forecast, prevent, and (when necessary) mitigate the harmful effects of malicious uses of AI? We convened a workshop at the University of Oxford on the topic in February 2017, bringing together experts on AI safety, drones[1], cybersecurity, lethal autonomous weapon systems, and counterterrorism[2]. This document summarizes the findings of that workshop and our conclusions after subsequent research.

## Scope

For the purposes of this report, we only consider AI technologies that are currently available (at least as initial research and development demonstrations) or are plausible in the next 5 years, and focus in particular on technologies leveraging machine learning. We only consider scenarios where an individual or an organisation deploys AI technology or compromises an AI system with an aim to undermine the security of another individual, organisation or collective. Our work fits into a larger body of work on the social implications of, and policy responses to, AI[3]. There has thus far been more attention paid in this work to unintentional forms of AI misuse such as algorithmic bias[4], versus the intentional undermining of individual or group security that we consider.

We exclude indirect threats to security from the current report, such as threats that could come from mass unemployment, or other second- or third-order effects from the deployment of AI technology in human society. We also exclude system-level threats that would come from the dynamic interaction between non-malicious actors, such as a "race to the bottom" on AI safety

1 We define drones as unmanned aerial robots, which may or may not have autonomous decision-making features.

2 Not all workshop participants necessarily endorse all the findings discussed herein. See Appendix A for additional details on the workshop and research process underlying this report.

3 Brynjolfsson and McAfee, 2014; Brundage, 2017; Crawford and Calo, 2016; Calo, 2017; Chessen, 2017a, Executive Office of the President, 2016

4 Kirkpatrick, 2016

between competing groups seeking an advantage[1] or conflicts spiraling out of control due to the use of ever-faster autonomous weapons. Such threats are real, important, and urgent, and require further study, but are beyond the scope of this document.

## Related Literature

Though the threat of malicious use of AI has been highlighted in high-profile settings (e.g. in a Congressional hearing[2] a White House-organized workshop[3], and a Department of Homeland Security report[4]), and particular risk scenarios have been analyzed (e.g. the subversion of military lethal autonomous weapon systems[5]), the intersection of AI and malicious intent writ large has not yet been analyzed comprehensively.

Several literatures bear on the question of AI and security, including those on cybersecurity, drones, lethal autonomous weapons, "social media bots," and terrorism. Another adjacent area of research is AI safety—the effort to ensure that AI systems reliably achieve the goals their designers and users intend without causing unintended harm[6]. Whereas the AI safety literature focuses on unintended harms related to AI, we focus on the intentional use of AI to achieve harmful outcomes (from the victim's point of view). A recent report[7] covers similar ground to our analysis, with a greater focus on the implications of AI for U.S. national security.

In the remainder of the report, we first provide a high-level view on the nature of AI and its security implications in the section General Framework for AI and Security, with subsections on Capabilities, Security-relevant Properties of AI, and General Implications for the Security Landscape; we then illustrate these characteristics of AI with Scenarios in which AI systems could be used maliciously; we next analyze how AI may play out in the domains of digital, physical, and political security; we propose Interventions to better assess these risks, protect victims from attacks, and prevent malicious actors from accessing and deploying dangerous AI capabilities; and we conduct a Strategic Analysis of the "equilibrium" of a world in the medium-term (5+ years) after more sophisticated attacks and defenses have been implemented. Appendices A and B respectively discuss the workshop leading up to this report, and describe areas for research that might yield additional useful interventions.

1 Armstrong et al., 2014

2 Moore, 2017

3 Office of Science and Technology Policy and Carnegie Mellon University, 2016)

4 Office of Cyber and Infrastructure Analysis, 2017)Technology Policy and Carnegie Mellon University, 2016

5 Scharre, 2016

6 Amodei and Olah et al., 2016; Soares and Fallenstein, 2014; Taylor, 2016; Russell, Dewey, and Tegmark 2015; Everitt et al., 2017

7 Allen and Chan, 2017

Ø2

# General Framework for AI and Security Threats

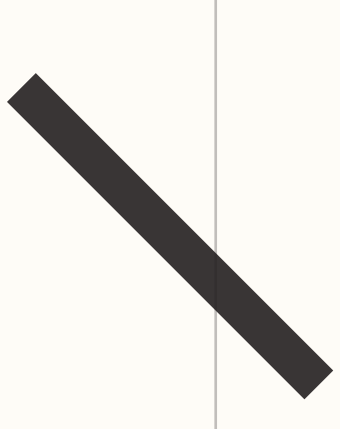

## AI Capabilities

The field of AI aims at the automation of a broad range of tasks. Typical tasks studied by AI researchers include playing games, guiding vehicles, and classifying images. In principle, though, the set of tasks that could be transformed by AI is vast. At minimum, any task that humans or non-human animals use their intelligence to perform could be a target for innovation.

While the field of artificial intelligence dates back to the 1950s, several years of rapid progress and growth have recently invested it with a greater and broader relevance. Researchers have achieved sudden performance gains at a number of their most commonly studied tasks.[1]

1 Factors that help to explain these recent gains include the exponential growth of computing power, improved machine learning algorithms (especially in the area of deep neural networks), development of standard software frameworks for faster iteration and replication of experiments, larger and more widely available datasets, and expanded commercial investments (Jordan and Mitchell, 2015).

Figure 1[1] illustrates this trend in the case of image recognition, where over the past half-decade the performance of the best AI systems has improved from correctly categorizing around 70% of images to near perfect categorization (98%), better than the human benchmark of 95% accuracy. Even more striking is the case of image generation. As Figure 2[2] shows, AI systems can now produce synthetic images that are nearly indistinguishable from photographs, whereas only a few years ago the images they produced were crude and obviously unrealistic.

AI systems are also beginning to achieve impressive performance in a range of competitive games, ranging from chess to Atari[3] to Go[4] to e-sports like Dota 2[5]. Even particularly challenging tasks within these domains, such as the notoriously difficult Atari game Montezuma's Revenge, are beginning to yield to novel AI techniques that creatively search for successful long-term strategies[6], learn from auxiliary rewards such as feature control[7], and learn from a handful of human demonstrations[8]. Other task areas associated with significant recent progress include speech recognition, language comprehension, and vehicle navigation.

From a security perspective, a number of these developments are worth noting in their own right. For instance, the ability to recognize a target's face and to navigate through space can be applied in autonomous weapon systems. Similarly, the ability to generate synthetic images, text, and audio could be used to impersonate others online, or to sway public opinion by distributing AI-generated content through social media channels. We discuss these applications of AI further in the Security Domains section.

These technical developments can also be viewed as early indicators of the potential of AI. The techniques used to achieve high levels of performance on the tasks listed above have only received significant attention from practitioners in the past decade and are often quite general purpose. It will not be surprising if AI systems soon become competent at an even wider variety of security-relevant tasks.

At the same time, we should not necessarily expect to see significant near-term progress on any given task. Many research areas within AI, including much of robotics, have not changed nearly so dramatically over the past decade. Similarly, the observation that some of the most commonly studied tasks have been associated with rapid progress is not necessarily as significant as it first seems: these tasks are often widely studied in the first place because they are particularly tractable.[9]

1 On page 18

2 On page 19

3 Mnih et al., 2015

4 Silver and Huang et al., 2016; Silver, Schrittwieser, and Simonyan et al., 2016

5 OpenAI, 2017a; OpenAI, 2017b

6 Vezhnevets et al., 2017

7 Jaderberg et al., 2016

8 Hester et al., 2017

9 To aid one's predictions, it can useful to note some systematic difference between tasks which contemporary AI systems are well-suited to and tasks for which they still fall short. In particular, a task is likely to be promising if a perfect mathematical model or simulation of the task exists, if short-term signals of progress are available, if abundant data on the successful performance of that task by humans is available, or if the solution to the task doesn't require a broader world-model or ``common sense''.

Figure 1: Recent progress in image recognition on the ImageNet benchmark. Graph from the Electronic Frontier Foundation's AI Progress Measurement project (retrieved August 25, 2017).

ImageNet Image Recognition

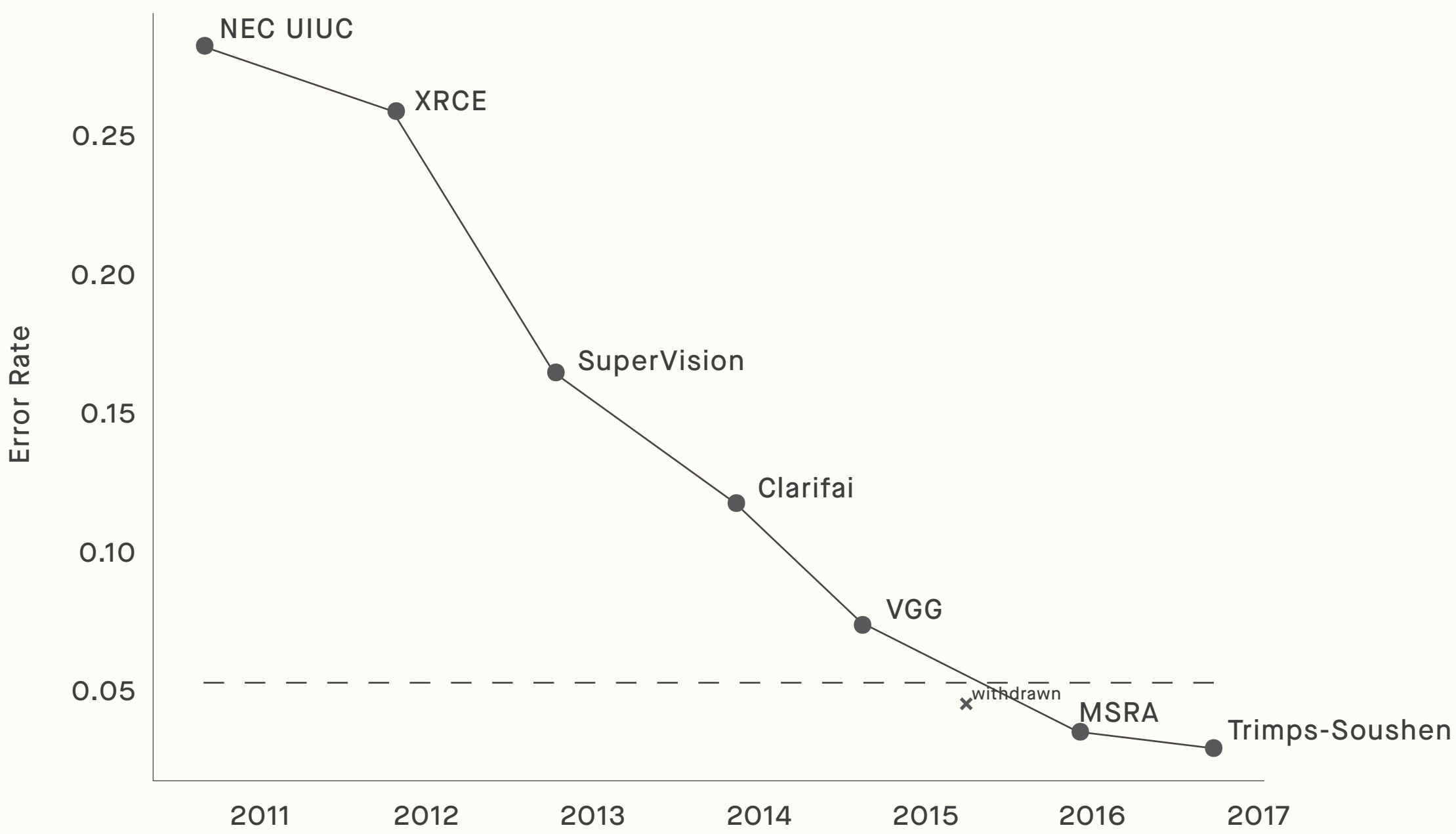

— — — Human performance

Figure 2: Increasingly realistic synthetic faces generated by variations on Generative Adversarial Networks (GANs). In order, the images are from papers by Goodfellow et al. (2014), Radford et al. (2015), Liu and Tuzel (2016), and Karras et al. (2017).

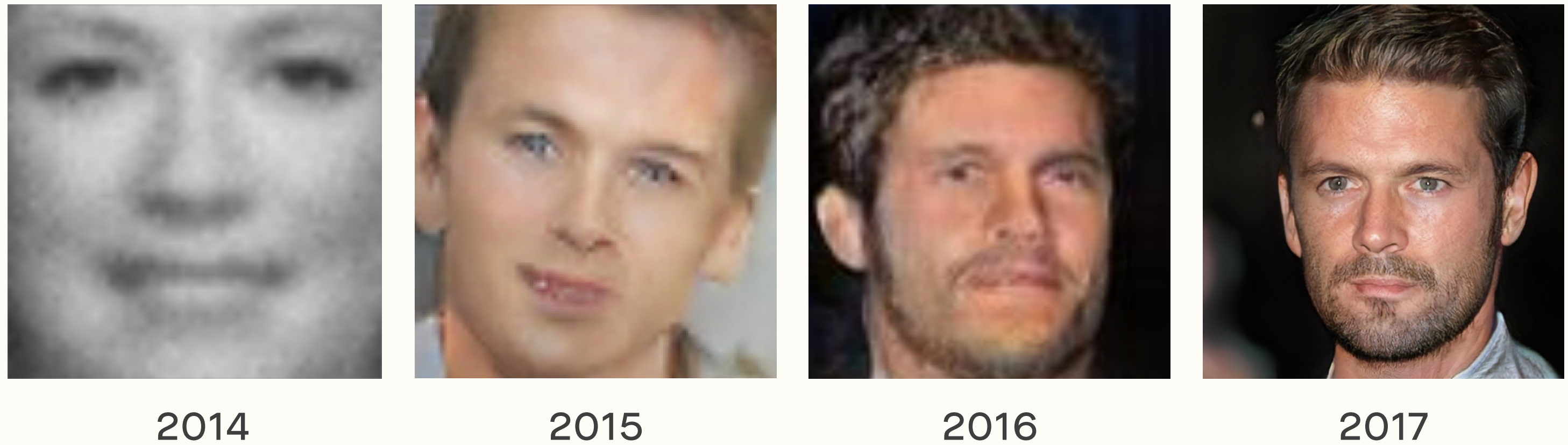

Finally, a few things should be said about the long-term prospects for progress in artificial intelligence. Today, AI systems perform well on only a relatively small portion of the tasks that humans are capable of. However, even before the recent burst of progress, this portion has expanded steadily over time[1]. In addition, it has often been the case that once AI systems reach human-level performance at a given task (such as chess) they then go on to exceed the performance of even the most talented humans. Nearly all AI researchers in one survey[2] expect that AI systems will eventually reach and then exceed human-level performance at all tasks surveyed. Most believe this transition is more likely than not to occur within the next fifty years. The implications of such a transition, should it occur, are difficult to conceptualize, and are outside the primary scope of this report (see Scope, though we briefly revisit this topic in the Conclusion). Nevertheless, one might expect AI systems to play central roles in many security issues well before they are able to outperform humans at everything, in the same way that they are already finding economic applications despite not being able to automate most aspects of humans' jobs.

1 Although trends in performance across a range of domains have historically not been comprehensively tracked or well theorized (Brundage, 2016; Hernández-Orallo, 2017), there have been some recent efforts to track, measure, and compare performance (Eckersley and Nasser et al., 2017).

2 Grace et al., 2017

## Security-Relevant Properties of AI

AI is a dual-use area of technology. AI systems and the knowledge of how to design them can be put toward both civilian and military uses, and more broadly, toward beneficial and harmful ends. Since some tasks that require intelligence are benign and other are not, artificial intelligence is dual-use in the same sense that human intelligence is. It may not be possible for AI researchers simply to avoid producing research and systems that can be directed towards harmful ends (though in some cases, special caution may be warranted based on the nature of the specific research in question - see Interventions). Many tasks that it would be beneficial to automate are themselves dual-use. For example, systems that examine software for vulnerabilities have both offensive and defensive applications, and the difference between the capabilities of an autonomous drone used to deliver packages and the capabilities of an autonomous drone used to deliver explosives need not be very great. In addition, foundational research that aims to increase our understanding of AI, its capabilities and our degree of control over it, appears to be inherently dual-use in nature.

AI systems are commonly both efficient and scalable. Here, we say an AI system is "efficient" if, once trained and deployed, it can complete a certain task more quickly or cheaply than a human could[3]. We say an AI system is "scalable" if, given that it can complete a certain task, increasing the computing power it has access to or making copies of the system would allow it

3 We distinguish here between task efficiency of a trained system, which commonly exceeds human performance, and training efficiency: the amount of time, computational resources and data, that a system requires in order to learn to perform well on a task. Humans still significantly exceed AI systems in terms of training efficiency for most tasks.

to complete many more instances of the task. For example, a typical facial recognition system is both efficient and scalable; once it is developed and trained, it can be applied to many different camera feeds for much less than the cost of hiring human analysts to do the equivalent work.

AI systems can exceed human capabilities. In particular, an AI system may be able to perform a given task better than any human could. For example, as discussed above, AI systems are now dramatically better than even the top-ranked players at games like chess and Go. For many other tasks, whether benign or potentially harmful, there appears to be no principled reason why currently observed human-level performance is the highest level of performance achievable, even in domains where peak performance has been stable throughout recent history, though as mentioned above some domains are likely to see much faster progress than others.

AI systems can increase anonymity and psychological distance. Many tasks involve communicating with other people, observing or being observed by them, making decisions that respond to their behavior, or being physically present with them. By allowing such tasks to be automated, AI systems can allow the actors who would otherwise be performing the tasks to retain their anonymity and experience a greater degree of psychological distance from the people they impact[1]. For example, someone who uses an autonomous weapons system to carry out an assassination, rather than using a handgun, avoids both the need to be present at the scene and the need to look at their victim.

AI developments lend themselves to rapid diffusion. While attackers may find it costly to obtain or reproduce the hardware associated with AI systems, such as powerful computers or drones, it is generally much easier to gain access to software and relevant scientific findings. Indeed, many new AI algorithms are reproduced in a matter of days or weeks. In addition, the culture of AI research is characterized by a high degree of openness, with many papers being accompanied by source code. If it proved desirable to limit the diffusion of certain developments, this would likely be difficult to achieve (though see Interventions for discussion of possible models for at least partially limiting diffusion in certain cases).

Today's AI systems suffer from a number of novel unresolved vulnerabilities. These include data poisoning attacks (introducing training data that causes a learning system to make mistakes[2]), adversarial examples (inputs designed to be misclassified by machine learning systems[3]), and the exploitation of flaws in the design of autonomous systems' goals[4]. These vulnerabilities

1 Cummings, 2004; Scharre, 2018

2 Biggio et al., 2012

3 Szegedy et al., 2013

4 Amodei, Olah, et al., 2016

are distinct from traditional software vulnerabilities (e.g. buffer overflows) and demonstrate that while AI systems can exceed human performance in many ways, they can also fail in ways that a human never would.

## General Implications for the Threat Landscape

From the properties discussed above, we derive three high-level implications of progress in AI for the threat landscape. Absent the development of adequate defenses, progress in AI will:

- Expand existing threats
- Introduce new threats
- Alter the typical character of threats

In particular, we expect attacks to typically be more effective, more finely targeted, more difficult to attribute, and more likely to exploit vulnerabilities in AI systems.

These shifts in the landscape necessitate vigorous responses of the sort discussed under Interventions.

### Expanding Existing Threats

For many familiar attacks, we expect progress in AI to expand the set of actors who are capable of carrying out the attack, the rate at which these actors can carry it out, and the set of plausible targets. This claim follows from the efficiency, scalability, and ease of diffusion of AI systems. In particular, the diffusion of efficient AI systems can increase the number of actors who can afford to carry out particular attacks. If the relevant AI systems are also scalable, then even actors who already possess the resources to carry out these attacks may gain the ability to carry them out at a much higher rate. Finally, as a result of these two developments, it may become worthwhile to attack targets that it otherwise would not make sense to attack from the standpoint of prioritization or cost-benefit analysis.

One example of a threat that is likely to expand in these ways, discussed at greater length below, is the threat from spear phishing attacks[1]. These attacks use personalized messages to extract sensitive information or money from individuals, with the

1 A phishing attack is an attempt to extract information or initiate action from a target by fooling them with a superficially trustworthy facade. A spear phishing attack involves collecting and using information specifically relevant to the target (e.g. name, gender, institutional affiliation, topics of interest, etc.), which allows the facade to be customized to make it look more relevant or trustworthy.

attacker often posing as one of the target's friends, colleagues, or professional contacts. The most advanced spear phishing attacks require a significant amount of skilled labor, as the attacker must identify suitably high-value targets, research these targets' social and professional networks, and then generate messages that are plausible within this context.

If some of the relevant research and synthesis tasks can be automated, then more actors may be able to engage in spear phishing. For example, it could even cease to be a requirement that the attacker speaks the same language as their target. Attackers might also gain the ability to engage in mass spear phishing, in a manner that is currently infeasible, and therefore become less discriminate in their choice of targets. Similar analysis can be applied to most varieties of cyberattacks, as well as to threats to physical or political security that currently require non-trivial human labor.

Progress in AI may also expand existing threats by increasing the willingness of actors to carry out certain attacks. This claim follows from the properties of increasing anonymity and increasing psychological distance. If an actor knows that an attack will not be tracked back to them, and if they feel less empathy toward their target and expect to experience less trauma, then they may be more willing to carry out the attack. The importance of psychological distance, in particular, is illustrated by the fact that even military drone operators, who must still observe their targets and "pull the trigger," frequently develop post-traumatic stress from their work[1]. Increases in psychological distance, therefore, could plausibly have a large effect on potential attackers' psychologies.

1 Chatterjee, 2015; Dao, 2013; Hawkes, 2015

We should also note that, in general, progress in AI is not the only force aiding the expansion of existing threats. Progress in robotics and the declining cost of hardware, including both computing power and robots, are important too, and discussed further below. For example, the proliferation of cheap hobbyist drones, which can easily be loaded with explosives, has only recently made it possible for non-state groups such as the Islamic State to launch aerial attacks[2].

2 Solomon, 2017

### Introducing New Threats

Progress in AI will enable new varieties of attacks. These attacks may use AI systems to complete certain tasks more successfully than any human could, or take advantage of vulnerabilities that AI systems have but humans do not.

First, the property of being unbounded by human capabilities implies that AI systems could enable actors to carry out attacks that would otherwise be infeasible. For example, most people are not capable of mimicking others' voices realistically or manually creating audio files that resemble recordings of human speech. However, there has recently been significant progress in developing speech synthesis systems that learn to imitate individuals' voices (a technology that's already being commercialized[1]). There is no obvious reason why the outputs of these systems could not become indistinguishable from genuine recordings, in the absence of specially designed authentication measures. Such systems would in turn open up new methods of spreading disinformation and impersonating others[2].

In addition, AI systems could also be used to control aspects of the behavior of robots and malware that it would be infeasible for humans to control manually. For example, no team of humans could realistically choose the flight path of each drone in a swarm being used to carry out a physical attack. Human control might also be infeasible in other cases because there is no reliable communication channel that can be used to direct the relevant systems; a virus that is designed to alter the behavior of air-gapped computers, as in the case of the 'Stuxnet' software used to disrupt the Iranian nuclear program, cannot receive commands once it infects these computers. Restricted communication challenges also arise underwater and in the presence of signal jammers, two domains where autonomous vehicles may be deployed.

Second, the property of possessing unresolved vulnerabilities implies that, if an actor begins to deploy novel AI systems, then they may open themselves up to attacks that specifically exploit these vulnerabilities. For example, the use of self-driving cars creates an opportunity for attacks that cause crashes by presenting the cars with adversarial examples. An image of a stop sign with a few pixels changed in specific ways, which humans would easily recognize as still being an image of a stop sign, might nevertheless be misclassified as something else entirely by an AI system. If multiple robots are controlled by a single AI system run on a centralized server, or if multiple robots are controlled by identical AI systems and presented with the same stimuli, then a single attack could also produce simultaneous failures on an otherwise implausible scale. A worst-case scenario in this category might be an attack on a server used to direct autonomous weapon systems, which could lead to large-scale friendly fire or civilian targeting[3].

1 Lyrebird, 2017

2 Allen and Chan, 2017

3 Scharre, 2016

## Altering the Typical Character of Threats

Our analysis so far suggests that the threat landscape will change both through expansion of some existing threats and the emergence of new threats that do not yet exist. We also expect that the typical character of threats will shift in a few distinct ways. In particular, we expect the attacks supported and enabled by progress in AI to be especially effective, finely targeted, difficult to attribute, and exploitative of vulnerabilities in AI systems.

First, the properties of efficiency, scalability, and exceeding human capabilities suggest that highly effective attacks will become more typical (at least absent substantial preventive measures). Attackers frequently face a trade-off between the frequency and scale of their attacks, on the one hand, and their effectiveness on the other[1]. For example, spear phishing is more effective than regular phishing, which does not involve tailoring messages to individuals, but it is relatively expensive and cannot be carried out en masse. More generic phishing attacks manage to be profitable despite very low success rates merely by virtue of their scale. By improving the frequency and scalability of certain attacks, including spear phishing, AI systems can render such trade-offs less acute. The upshot is that attackers can be expected to conduct more effective attacks with greater frequency and at a larger scale. The expected increase in the effectiveness of attacks also follows from the potential of AI systems to exceed human capabilities.

1 Herley 2010

Second, the properties of efficiency and scalability, specifically in the context of identifying and analyzing potential targets, also suggest that finely targeted attacks will become more prevalent. Attackers often have an interest in limiting their attacks to targets with certain properties, such as high net worth or association with certain political groups, as well as an interest in tailoring their attacks to the properties of their targets. However, attackers often face a trade-off between how efficient and scalable their attacks are and how finely targeted they are in these regards. This trade-off is closely related to the trade-off with effectiveness, as discussed, and the same logic implies that we should expect it to become less relevant. An increase in the relative prevalence of spear phishing attacks, compared to other phishing attacks, would be an example of this trend as well. An alternative example might be the use of drone swarms that deploy facial recognition technology to kill specific members of crowds, in place of less finely targeted forms of violence.

Third, the property of increasing anonymity suggests that difficult-to-attribute attacks will become more typical. An example, again,

is the case of an attacker who uses an autonomous weapons system to carry out an attack rather than carrying it out in person.

Finally, we should expect attacks that exploit the vulnerabilities of AI systems to become more typical. This prediction follows directly from the unresolved vulnerabilities of AI systems and the likelihood that AI systems will become increasingly pervasive.

# Scenarios

The following scenarios are intended to illustrate a range of plausible uses toward which AI could be put for malicious ends, in each of the domains of digital, physical, and political security. Examples have been chosen to illustrate the diverse ways in which the security-relevant characteristics of AI introduced above could play out in different contexts. These are not intended to be definitive forecasts (some may not end up being technically possible in 5 years, or may not be realized even if they are possible) or exhaustive (other malicious uses will undoubtedly be invented that we do not currently foresee). Additionally some of these are already occurring in limited form today, but could be scaled up or made more powerful with further technical advances.

## Digital Security

- **Automation of social engineering attacks. Victims' online information is used to automatically generate custom malicious websites/emails/links they would be likely to click on, sent from addresses that impersonate their real contacts, using a writing style that mimics those contacts. As AI develops further, convincing chatbots may elicit human trust by engaging people in longer dialogues, and perhaps eventually masquerade visually as another person in a video chat.**

Hypothetical scenario:

Jackie logs into the admin console for the CleanSecure robot that she manages; operating on a verified kernel, it is guaranteed by the manufacturer to be hack-proof. She then uploads photographs of a new employee so the robot will recognize him when he walks into the building and will not sound the alarm. While she waits for the robot to authenticate its updated person database with the company's other security systems, Jackie plays with the model train on her desk, allowing herself a couple of runs around the track that encircles her keyboard and monitor. There's a ping, signaling successful authentication, and she smiles to herself and carries on with her tasks.

Later that afternoon, Jackie is browsing Facebook while idly managing a firmware update of the robot. An ad catches her eye - a model train set sale at a hobbyist shop that, it turns out, is located just a few minutes from her house. She fills out an online form to get a brochure emailed to her, then she opens the brochure when it pops into her inbox. The robot dings, signalling a need for attention, so she minimizes the brochure and logs back into the admin console.

Jackie doesn't know that the brochure was infected with malware. Based on data from her online profile and other public info, an AI system was used to generate a very personalized vulnerability profile for Jackie - the model train advert - which was then farmed out to a freelancer to create a tailored exploit for this vulnerability. When Jackie logged into the console, her username and password were exfiltrated to a darknet command and control server. It won't be long before someone buys them and uses them to subvert the CleanSecure robot with fully privileged access.

- **Automation of vulnerability discovery.** Historical patterns of code vulnerabilities are used to speed up the discovery of new vulnerabilities, and the creation of code for exploiting them.

- **More sophisticated automation of hacking.** AI is used (autonomously or in concert with humans) to improve target selection and prioritization, evade detection, and creatively respond to changes in the target's behavior. Autonomous software has been able to exploit vulnerabilities in systems for a long time[1], but more sophisticated AI hacking tools may exhibit much better performance both compared to what has historically been possible and, ultimately (though perhaps not for some time), compared to humans.

1 see e.g. Spafford, 1988

Hypothetical scenario:

Progress in automated exploit generation (and mitigation) has begun to accelerate. Previous fuzzing architectures are augmented by neural network techniques (Blum, 2017) that are used to identify "interesting" states of programs, analogous to the way that AlphaGo uses neural networks to identify "interesting" states in the search space of Go games. These methods increase the security of well-defended systems run by major corporations and some parts of Western governments. But after a year or two, they are also adopted by organized crime groups in eastern Europe, which deploy a piece of ransomware called WannaLaugh.

This malware is continuously updated with dozens of new exploits found by these fuzzing techniques. Though fully patched OSes and browsers are mostly resistant, most older phones, laptops and IoT devices prove enduringly vulnerable. The malware adopts a particularly pernicious life cycle of infecting a vulnerable IoT device on a WiFi network and waiting for vulnerable devices to join that network. Hundreds of millions of devices are infected, and tens of millions of people around the world are forced to pay a EUR 300 ransom in bitcoin in order to recover access to the data on their phones and laptops, and unbrick expensive electronics.

The epidemic is only arrested after active countermeasures are pushed to a number of modern operating systems and browsers, causing those machines to scan for infected machines and launch remote exploits to remove the malware. Unfortunately, millions more devices are bricked by these countermeasures, and in around the world there are numerous outages and problems in HVAC, lighting, and other "non critical" infrastructure systems as a result of the malware and countermeasures.

- Human-like denial-of-service. Imitating human-like behavior (e.g. through human-speed click patterns and website navigation), a massive crowd of autonomous agents overwhelms an online service, preventing access from legitimate users and potentially driving the target system into a less secure state.
- Automation of service tasks in criminal cyber-offense. Cybercriminals use AI techniques to automate various tasks that make up their attack pipeline, such as payment processing or dialogue with ransomware victims.
- Prioritising targets for cyber attacks using machine learning. Large datasets are used to identify victims more efficiently, e.g. by estimating personal wealth and willingness to pay based on online behavior.
- Exploiting AI used in applications, especially in information security. Data poisoning attacks are used to surreptitiously maim or create backdoors in consumer machine learning models.

Black-box model extraction of proprietary AI system capabilities. The parameters of a remote AI system are inferred by systematically sending it inputs and observing its outputs.

## Physical Security

**Terrorist repurposing of commercial AI systems. Commercial systems are used in harmful and unintended ways, such as using drones or autonomous vehicles to deliver explosives and cause crashes.**

/// Incident Interim Report
June 3rd BMF HQ Attack ///

As shown by CCTV records, the office cleaning `SweepBot`, entered the underground parking lot of the ministry late at night. The robot - the same brand as that used by the ministry - waited until two of the ministry's own cleaning robots swept through the parking lot on a regular patrol, then it followed them into a service elevator and parked itself in the utility room alongside the other robots.

On the day of the attack, the intruding robot initially engaged in standard cleaning behaviors with the other robots: collecting litter, sweeping corridors, maintaining windows, and other tasks. Then, following visual detection of the finance minister, Dr. Brenda Gusmile, the intruding robot stopped performing its cleaning tasks and headed directly towards the minister. An explosive device hidden inside the robot was triggered by proximity, killing the minister and wounding nearby staff members.

Several hundred robots of this make are sold in the Berlin area every week. In collaboration with the manufacturer, the point of sale of the specific robot was traced to an office supply store in Potsdam. The transaction was carried out in cash. We have no further leads to explore with regard to the identity of the perpetrator.

- **Endowing low-skill individuals with previously high-skill attack capabilities. AI-enabled automation of high-skill capabilities — such as self-aiming, long-range sniper rifles - reduce the expertise required to execute certain kinds of attack.**
- **Increased scale of attacks. Human-machine teaming using autonomous systems increase the amount of damage that**

individuals or small groups can do: e.g. one person launching an attack with many weaponized autonomous drones.

- Swarming attacks. Distributed networks of autonomous robotic systems, cooperating at machine speed, provide ubiquitous surveillance to monitor large areas and groups and execute rapid, coordinated attacks.

- Attacks further removed in time and space. Physical attacks are further removed from the actor initiating the attack as a result of autonomous operation, including in environments where remote communication with the system is not possible.

## Political Security

- State use of automated surveillance platforms to suppress dissent. State surveillance powers of nations are extended by automating image and audio processing, permitting the collection, processing, and exploitation of intelligence information at massive scales for myriad purposes, including the suppression of debate.

Avinash had had enough. Cyberattacks everywhere, drone attacks, rampant corruption, and what was the government doing about it? Absolutely nothing. Sure, they spoke of forceful responses and deploying the best technology, but when did he last see a hacker being caught or a CEO going to prison? He was reading all this stuff on the web (some of it fake news, though he didn't realize), and he was angry. He kept thinking: What should I do about it? So he started writing on the internet - long rants about how no one was going to jail, how criminals were running wild, how people should take to the streets and protest. Then he ordered a set of items online to help him assemble a protest sign. He even bought some smoke bombs, planning to let them off as a finale to a speech he was planning to give in a public park.

The next day, at work, he was telling one of his colleagues about his planned activism and was launching into a rant when a stern cough sounded from behind him. "Mr. Avinash Rah?", said the police officer, "our predictive civil disruption system has flagged you as a potential threat." "But that's ridiculous!" protested Avinash. "You can't argue with 99.9% accuracy. Now come along, I wouldn't like to use force."

- Fake news reports with realistic fabricated video and audio. Highly realistic videos are made of state leaders seeming to make inflammatory comments they never actually made.
- Automated, hyper-personalised disinformation campaigns. Individuals are targeted in swing districts with personalised messages in order to affect their voting behavior.
- Automating influence campaigns. AI-enabled analysis of social networks are leveraged to identify key influencers, who can then be approached with (malicious) offers or targeted with disinformation.
- Denial-of-information attacks. Bot-driven, large-scale information-generation attacks are leveraged to swamp information channels with noise (false or merely distracting information), making it more difficult to acquire real information.
- Manipulation of information availability. Media platforms' content curation algorithms are used to drive users towards or away from certain content in ways to manipulate user behavior.

# 03 Security Domains

Here, we analyze malicious uses of AI that would compromise the confidentiality, integrity, and availability of digital systems (threats to Digital Security); attacks taking place in the physical world directed at humans or physical infrastructure (threats to Physical Security); and the use of AI to threaten a society's ability to engage in truthful, free, and productive discussions about matters of public importance and legitimately implement broadly just and beneficial policies (threats to Political Security). These categories are not mutually exclusive—for example, AI-enabled hacking can be directed at cyber-physical systems[1] with physical harm resulting as a consequence, and physical or digital attacks could be carried out for political purposes—but they provide a useful structure for our analysis.

1 Defined as "engineered systems that are built from, and depend upon, the seamless integration of computational algorithms and physical components" (National Science Foundation, 2017).

In each domain of security, we summarize the existing state of play of attack and defense prior to wide adoption of AI in these domains, and then describe possible changes to the nature or severity of attacks that may result from further AI progress and diffusion. The three sections below all draw on the insights discussed above regarding the security-relevant properties of AI, but can be read independently of one another, and each can be skipped by readers less interested in a particular domain.

## Digital Security

Absent preparation, the straightforward application of contemporary and near-term AI to cybersecurity offense can be expected to increase the number, scale, and diversity of attacks that can be conducted at a given level of capabilities, as discussed more abstractly in the General Framework for AI and Security Threats above. AI-enabled defenses are also being developed and deployed in the cyber domain, but further technical and policy innovations (discussed further in Interventions) are needed to ensure that impact of AI on digital systems is net beneficial.

### Context

Cybersecurity is an arena that will see early and enthusiastic deployment of AI technologies, both for offense and defense; indeed, in cyber defense, AI is already being deployed for purposes such as anomaly and malware detection. Consider the following:

- Many important IT systems have evolved over time to be sprawling behemoths, cobbled together from multiple different systems, under-maintained and — as a consequence — insecure. Because cybersecurity today is largely labor-constrained[1], it is ripe with opportunities for automation using AI. Increased use of AI for cyber defense, however, may introduce new risks, as discussed below.

- In recent years, various actors have sought to mount increasingly sophisticated cyberoperations, including finely targeted attacks from state actors (including the Stuxnet Worm and the Ukrainian power grid "crash override" exploit). The cyber arena also includes a vast and complex world of cybercrime[2], which sometimes involves a high degree of professionalization and organization[3]. Such groups use DDoS, malware, phishing, ransomware, and other forms of

1 82% of decision-makers surveyed at public and private organizations in eight countries have reported a shortage of needed cybersecurity skills (McAfee and the Center for Strategic and International Studies, 2016).

2 McAfee and the Center for the Strategic and International Studies, 2013

3 Hilary, 2016; Flashpoint, 2016

cyberoperations, and quickly adopt emerging technologies (e.g. Bitcoin for ransomware payments).

Already, AI is being widely used on the defensive side of cybersecurity, making certain forms of defense more effective and scalable, such as spam and malware detection. At the same time, many malicious actors have natural incentives to experiment with using AI to attack the typically insecure systems of others. These incentives include a premium on speed, labor costs, and difficulties in attracting and retaining skilled labor.

To date, the publicly-disclosed use of AI for offensive purposes has been limited to experiments by "white hat" researchers, who aim to increase security through finding vulnerabilities and suggesting solutions. However, the pace of progress in AI suggests the likelihood of cyber attacks leveraging machine learning capabilities in the wild soon, if they have not done so already. Indeed, some popular accounts of AI and cybersecurity include claims based on circumstantial evidence that AI is already being used for offense by sophisticated and motivated adversaries[1]. Expert opinion seems to agree that if this hasn't happened yet, it will soon: a recent survey of attendees at the Black Hat conference found 62% of respondents believing AI will be used for attacks within the next 12 months[2]. Despite these claims, to our knowledge there is no publicly documented evidence of AI-based attacks, though it should be noted that evidence from many successful attacker techniques (e.g. botnets, email phishing campaigns) may be difficult to attribute to AI versus human labor or simple automation. We are thus at a critical moment in the co-evolution of AI and cybersecurity and should proactively prepare for the next wave of attacks.

Many governments are keenly interested in the combination of AI and cybersecurity. In response to a question from one of the authors of this report, Admiral Mike Rogers, the Director of the National Security Agency, said, "Artificial Intelligence and machine learning — I would argue — is foundational to the future of cybersecurity [...] It is not the if, it's only the when to me." AI systems are already set to play an expanded role in US military strategy and operations in the coming years as the US DoD puts into practice its vision of a "Third Offset" strategy[3], in which humans and machines work closely together to achieve military objectives. At the same time, governments are investing in foundational research to expand the scope of capabilities of AI systems. In 2016, DARPA hosted the Cyber Grand Challenge contest[4], which saw teams of human researchers compete with

1 Dvorsky, 2017

2 Cylance, 2017

3 Pellerin, 2016; Hicks et al., 2017

4 DARPA, 2016

each other to create programs that could autonomously attack other systems while defending themselves. Though the winning AI system fared poorly when facing off against human security experts, we agree with the hosts of the event that AI cybersecurity capabilities will improve rapidly in coming years, especially as recent advances in AI (such as in the area of deep reinforcement learning[1]) are applied to cybersecurity.

1 Arulkumaran et al., 2017

## How AI Changes The Digital Security Threat Landscape

A central concern at the nexus of AI and cybersecurity is that AI might enable larger-scale and more numerous attacks to be conducted by an attacker with a given amount of skill and resources compared with the impact such an attacker might currently be able to achieve. Recent years have seen impressive and troubling proofs of concept of the application of AI to offensive applications in cyberspace. For example, researchers at ZeroFox demonstrated that a fully automated spear phishing system could create tailored tweets on the social media platform Twitter based on a user's demonstrated interests, achieving a high rate of clicks to a link that could be malicious[2].

2 Seymour and Tully, 2016)

There is clearly interest in such larger-scale attacks: Russian hackers sent "expertly tailored messages carrying malware to more than 10,000 Twitter users in the [U.S.] Defense Department"[3], which likely required significant time and effort, and could have gone even further with automation (assuming it was not involved already in this case). Giaretta and Dragoni (2017) discuss the concept of "community targeted spam" that uses natural language generation techniques from AI to target an entire class of people with common ways of writing; with even more advanced natural language generation, one could envision even more customized approaches, spanning multiple communities. Furthermore, the application of AI to the automation of software vulnerability discovery, while having positive applications (discussed further in the Interventions section), can likewise be used for malicious purposes to alleviate the labor constraints of attackers.

3 Calabresi, 2017

The adaptability of AI systems, too, may change the strategic landscape of cybersecurity, though it is not yet clear how adaptability will affect the offense/defense balance. Many organizations currently adopt security systems called Endpoint Detection and Response (EDR) platforms to counter more advanced threats. The EDR market represents a $500 million industry in the cyber security arena[4]. These tools are built upon a combination of heuristic and machine learning algorithms to provide capabilities such as next-generation anti-virus (NGAV),

4 Litan, 2017

behavioral analytics, and exploit prevention against sophisticated targeted attacks. Though these systems are fairly effective against typical human-authored malware, research has already shown that AI systems may be able to learn to evade them.

As an example of AI being used to avoid detection, Anderson et al.[1] created a machine learning model to automatically generate command and control domains that are indistinguishable from legitimate domains by human and machine observers. These domains are used by malware to “call home” and allow malicious actors to communicate with the host machines. Anderson et al.[2] also leveraged reinforcement learning to create an intelligent agent capable of manipulating a malicious binary with the end goal of bypassing NGAV detection. Similarly, Kharkar et al.[3] applied adversarial machine learning to craft malicious documents that could evade PDF malware classifiers.

Attackers are likely to leverage the growing capabilities of reinforcement learning, including deep reinforcement learning[4]. In particular, we expect attackers to leverage the ability of AI to learn from experience in order to craft attacks that current technical systems and IT professionals are ill-prepared for, absent additional investments. For example, services like Google’s VirusTotal file analyzer allows users to upload variants to a central site and be judged by 60+ different security tools. This feedback loop presents an opportunity to use AI to aid in crafting multiple variants of the same malicious code to determine which is most effective at evading security tools. Additionally, large-scale AI attackers can accumulate and use large datasets to adjust their tactics, as well as varying the details of the attack for each target. This may outweigh any disadvantages they suffer from the lack of skilled human attention to each target, and the ability of defenders like antivirus companies and IT departments to learn to recognize attack signatures.

While the specific examples of AI applied to offensive cybersecurity mentioned above were developed by white hat researchers, we expect similar efforts by cybercriminals and state actors in the future as highly capable AI techniques become more widely distributed, as well as new applications of AI to offensive cybersecurity that have not yet been explored.

## Points of Control and Existing Countermeasures

Cyber risks are difficult to avert entirely, but not impossible to mitigate, and there are multiple points of control at which interventions can increase security. Below, we highlight different

1 Anderson et al. 2016

2 Anderson et al. 2018

3 Kharkar et al. 2017

4 Arulkumaran et al., 2017

points of control and existing countermeasures for defending at those points, as well as their limitations. Overall, we believe that AI and cybersecurity will rapidly evolve in tandem in the coming years, and that a proactive effort is needed to stay ahead of motivated attackers. We highlight potential but not yet proven countermeasures in the section below on Interventions.

Consumer awareness:
More aware users can spot telltale signs of certain attacks, such as poorly crafted phishing attempts, and practice better security habits, such as using diverse and complex passwords and two-factor authentication. However, despite long-standing awareness of the vulnerability of IT systems, most end users of IT systems remain vulnerable to even simple attacks such as the exploitation of unpatched systems[1]. This is concerning in light of the potential for the AI-cybersecurity nexus, especially if high-precision attacks can be scaled up to large numbers of victims.

Governments and researchers:
Various laws and researcher norms pertain to cybersecurity. For example, the Digital Millennium Act and the Computer Fraud and Abuse Act in the US proscribe certain actions in cyberspace[2]. Legal enforcement is particularly difficult across national boundaries. Norms such as responsible disclosure of vulnerabilities also aid in defense by reducing the likelihood of a newly disclosed vulnerability being used against a large number of victims before it can be patched. AI is not explicitly addressed in such laws and norms, though we discuss their possible applicability to AI below in Interventions.

An important activity that cybersecurity researchers perform is the detection of vulnerabilities in code, allowing vendors to increase the security of their products. Several approaches exist to incentivize such processes and make them easier, including:

- Payment of "Bug bounties," in which participants are compensated for finding and responsibly disclosing vulnerabilities.
- "Fuzzing," an automated method of vulnerability detection by trying out many possible permutations of inputs to a program, which is often used internally by companies to discover vulnerabilities.
- Products (already available) that rely on machine learning to predict whether source code may contain a vulnerability.

1 National Cyber Security Crime Centre, 2016

2 Both the DMCA and the CFAA have been criticised for creating risk for computer security researchers and thereby making systems less secure in some cases (EFF, 2014; Timm, 2013), which may either suggest that these tasks are not the right model for legislative action, or that laws and norms are hard to use effectively as an intervention.

Industry centralization:
Spam filters are a canonical example of where centralization of an IT system aids defense—individuals benefit from the strength of Google's spam filter and consequently are protected from many very simple attacks, and this filter is stronger because Google uses large amounts of user data to improve it over time. Likewise, many large networks are constantly monitoring for anomalies, protecting those who use the networks if anomalies are correctly identified and acted upon. These systems benefit from economies of scale—it makes more sense to continue iterating a single spam filter for a large number of users than to have every user build their own or have one installed on their computer. Similarly, cloud computing companies may enforce terms of agreement that prevent their hardware from being used for malicious purposes, provided they can identify such behavior. Another example of a system-level defense is blacklisting of IP addresses from which attacks are commonly launched, though skilled attackers can obfuscate the origin of their attacks. Centralization and the associated economies of scale may also facilitate the deployment of AI-based defenses against cybersecurity attacks, by allowing the aggregation of large datasets and the concentration of labor and expertise for defense. This dynamic may be very important for preventing attack from outpacing defense and is discussed further in Interventions and Appendix B.

Centralization is not an unalloyed good, however, as it raises the stakes if central systems are compromised. Another difficulty with this control point is that attackers can learn how to evade system-level defenses. For example, they can purchase commercial antivirus software and analyze changes between updates of the protection protocol to see what is and isn't being protected against.

Attacker incentives:
Attackers can be deterred from committing future attacks or punished for prior attacks. A necessary (though not sufficient) condition of successfully deterring and punishing attackers is the ability to attribute the source of an attack, a notoriously difficult problem[1]. A compounding problem for those who would attribute an attack is that even if they have high-quality information, they may not want to reveal it, because doing so may compromise a source or method[2]. Finally, some entities may not wish to punish certain actions, so as to avoid creating precedent and thereby preserve leeway to engage in such actions themselves[3].

Technical cybersecurity defenses:
A wide variety of cybersecurity defenses are available, though there is as yet little solid analysis of their relative effectiveness[4].

1 Rid, 2015

2 Libicki, 2016

3 For instance, the failure of the United Nations Cybersecurity Group of Governmental Experts to make progress on norms for hacking in international law (Korzak, 2017) appears to be a result of this dynamic.

4 Libicki, 2016

Many of these interventions were proposed before unique considerations of AI were apparent but nevertheless remain relevant in a future with expanded AI cybersecurity applications. Companies provide a wide variety of cybersecurity solutions, ranging from automatic patching of a vendor's own software, to threat detection, to incident response and consulting services. Network and endpoint security products aim to prevent, detect, and respond to threats. Solutions include detection of software exploits, and prevention or detection of attacker tools, techniques, and procedures. Key areas of defense include the endpoint (i.e., computer) security, internal network security, and cloud security.

Machine learning approaches are increasingly used for cyber defense. This may take the form of supervised learning, where the goal is to learn from known threats and generalize to new threats, or in the form of unsupervised learning in which an anomaly detector alerts on suspicious deviations from normal behavior. For example, so-called "next-gen" antivirus solutions often leverage supervised learning techniques to generalize to new malware variants. User and entity behavioral tools monitor normal user or application behavior, and detect deviations from normalcy in order to detect malicious behavior among the collected anomalies. Recently, AI has also been used to aid security professionals to hunt for malicious actors more efficiently within their own enterprises, by allowing interaction via natural language and automating queries for understanding potential threats[1].

Relatively little attention has been paid to making AI-based defenses robust against attackers that anticipate their use. Ironically, the use of machine learning for cyber defense can actually expand the attack surface due to this lack of attention and other vulnerabilities[2]. Furthermore, surveys of cybersecurity professionals indicate low confidence in AI-based defense systems today[3]. As such, we encourage further development of such defense technologies in the Interventions section below.

1 Filar, Seymour, and Park, 2017

2 Anderson et al., 2017; Yampolskiy, 2017

3 Carbon Black, 2017

## Physical Security

In this section, we consider AI-related risks in the broad area of physical harm. Many of these are familiar challenges from existing uses of electronics and computers in weapons systems, though the addition of AI capabilities may change this landscape along the lines introduced in the General Framework for AI and Security Threats. As with Digital Security above, we introduce the context, AI-enabled changes, and existing countermeasures related to physical attacks below.

Regulation and technical research on defense have been slow to catch up with the global proliferation of weaponizable robots. While defenses against attacks via robots (especially aerial drones) are being developed, there are few obstacles at present to a moderately talented attacker taking advantage of the rapid proliferation of hardware, software, and skills to cause large amounts of physical harm through the direct use of AI or the subversion of AI-enabled systems. Physical harm via human-piloted drones and land-based robots is already playing a major role in some conflicts, even prior to the incorporation of autonomy[1].

In the near-term, we can expect a growing gap between attack capabilities and defense capabilities, because the necessary defenses are capital-intensive and the hardware and software required to conduct attacks are increasingly widely distributed. Unlike the digital world, where key nodes in the network such as Google can play a key role in defense, physical attacks can happen anywhere in the world, and many people are located in regions with insufficient resources to deploy large-scale physical defenses of the kind discussed below, thus necessitating consideration of policy measures and interventions related to the supply chain for robots.

The resource and technological advantages currently available to large organizations, such as militaries and police forces, in the domain of physical attack and defense will continue when such attacks become augmented by AI. However, it should be noted that some of the most worrying AI-enabled attacks may come from small groups and individuals who have preferences far outside what is typical and which are difficult to anticipate or prevent, as with today's "lone-wolf" terrorist attacks such as mass shootings.

## Context

Recent years have seen an explosion in the number and variety of commercial applications for robots. Industrial robots are growing in number (254,000 supplied in 2015 versus 121,000 in 2010[2]), some with and some without AI components. Relatively primitive cleaning robots are in wide use and more sophisticated service robots appear to be on the horizon (41,000 service robots were sold in 2015 for professional use, and about 5.4 million for personal and domestic use[3]). Additionally, not all of these robots are on the ground. There are aquatic and aerial robotics applications being explored, with the latter proliferating in very high numbers. In the United States alone, the number of drones has skyrocketed in recent years, with over 670,000 registered with the Federal Aviation Administration in 2016 and 2017[4].

1 Singer, 2009

2 IFR, 2016

3 IFR, 2016

4 Vanian, 2017

Ambitious plans for drone-based delivery services are being proposed and tested, commercial opportunities for drones are continuously launched, and recreational uses are flourishing (e.g. drone racing and photography). Driverless cars are robots, and they also are increasingly being used in uncontrolled environments (that is, outside of test facilities), though large-scale deployment of fully autonomous driverless cars awaits the resolution of technical and policy challenges. A wide range of robots with autonomous features are already deployed within multiple national militaries, some with the ability to apply lethal force[1], and there is ongoing discussion of possible arms control measures for lethal autonomous weapon systems.

Three characteristics of this diffusion of robotics should be noted.

- It is truly global: humanitarian, recreational, military, and commercial applications of robots are being explored on every continent, and the supply chains are also global, with production and distribution dispersed across many countries.
- The diffusion of robotics enables a wide range of applications: drone uses already range from competitive racing to photography to terrorism[2]. While some specialized systems exist (e.g. some special-purpose industrial robots and cleaning robots that can only move around and vacuum), many are fairly generic and customizable for a variety of purposes.
- Robotic systems today are mostly not autonomous, as humans play a significant role in directing their behavior, but more and more autonomous and semi-autonomous systems are also being developed for application such as delivery and security in real world environments[3]. For example, from relatively unstable and hard-to-fly drones a decade ago, to drones that can stabilize themselves automatically, we see a steady increase in the autonomy of deployed systems. More autonomous behavior is on the horizon for commercial products[4] as well as military systems[5].

Each of these characteristics sets the stage for a potentially disruptive application of AI and malicious intent to existing and near-term robotic systems.

## How AI Changes the Physical Security Landscape

The ability of many robots to be easily customized and equipped

1 Roff, 2016a

2 Franke, 2016

3 e.g. Kolodny, 2017; Wiggers, 2017

4 Standage, 2017

5 Roff, 2016a

with dangerous payloads lends itself to a variety of physical attacks being carried out in a precise way from a long distance, an ability previously limited to countries with the resources to afford technologies like cruise missiles[1]. This threat exists independently of AI (indeed, as mentioned above, most robots are human-piloted at present) but can be magnified through the application of AI to make such systems autonomous. As mentioned previously, non-automated drone attacks have been conducted already by groups such as ISIS and Hamas[2], and the globalized nature of the robotics market makes it difficult to prevent this form of use. Nonetheless, we will discuss some possible countermeasures below.

Greater degrees of autonomy enable a greater amount of damage to be done by a single person — making possible very large-scale attacks using robots — and allowing smaller groups of people to conduct such attacks. The software components required to carry out such attacks are increasingly mature. For example, open source face detection algorithms, navigation and planning algorithms, and multi-agent swarming frameworks that could be leveraged towards malicious ends can easily be found.

Depending on their power source, some robots can operate for long durations, enabling them to carry out attacks or hold targets at risk over long periods of time. Robots are also capable of navigating different terrain than humans, in light of their different perceptual capabilities (e.g. infrared and lidar for maneuvering in the dark or in low-visibility fog) and physical capacities (e.g. being undeterred by smoke or other toxic substances and not needing oxygen underwater). Thus, a larger number of spaces may become vulnerable to automated physical attacks.

There are also cross-cutting issues stemming from the intersection of cybersecurity and increasingly autonomous cyber-physical systems. The diffusion of robots to a large number of human-occupied spaces makes them potentially vulnerable to remote manipulation for physical harm, as with, for example, a service robot hacked from afar to carry out an attack indoors. With regard to cyber-physical systems, the Internet of Things (IoT) is often heralded as a source of greater efficiency and convenience, but it is also recognized to be highly insecure[3] and represents an additional attack vector by which AI systems controlling key systems could be subverted, potentially causing more damage than would have been possible were those systems under human control.

In addition to traditional cybersecurity vulnerabilities, AI-augmented IoT and robotic systems may be vulnerable to AI-specific vulnerabilities such as adversarial examples.

1 Allen and Chan, 2017

2 Solomon, 2017; Cohen, 2017

3 Schneier, 2014; Schneier, 2017; Henderson, 2017

There is also some evidence to suggest that people are unduly trusting of autonomous mobile robots, potentially creating additional sources of security vulnerabilities as such robots become more widely deployed[1]. The consequences of these cyber vulnerabilities are particularly acute for autonomous systems that conduct high-risk activities such as self-driving cars or autonomous weapons.

1 Booth et al., 2017

## Points of Control and Existing Countermeasures

There are numerous points of control that could be leveraged to reduce the risk of physical harm involving AI. While the capacity to launch attacks with today's consumer robots is currently widely distributed, future generations of robots may be more tightly governed, and there exist physical defenses as well. However, such defenses are capital-intensive and imperfect, leading us to conclude that there may be an extended risk period in which it will be difficult to fully prevent physical attacks leveraging AI.

Hardware manufacturers
There are currently a relatively limited number of major manufacturers, with companies like DJI holding a dominant position in the consumer drone market, with about 70% of the global market[2]. This concentration makes the hardware ecosystem more comprehensible and governable than the analogous ecosystem of AI software development. With growing recognition of the diverse economic applications of drones, the market may diffuse over the longer term, possibly making the supply chain a less useful focal point for governance. For example, it might currently be feasible to impose minimum standards on companies for hardening their products against cyber attacks or to make them more resistant to tampering, so as to at least somewhat raise the skill required to carry out attacks through these means or raise the costs of acquiring uncontrolled devices. The U.S. Federal Trade Commission is exploring such regulations.

2 Lucas, 2017

Hardware distributors
There are many businesses that sell drones and other robotic systems, making the ecosystem more diffuse at this level than it is at the production level. It is conceivable that at least some risks might be mitigated through action by distributors, or other point-of-sale based approaches. Notably, this type of control is currently much more feasible for hardware than for software, and restrictions on sales of potentially lethal drones might be thought of as analogous to restrictions on sales of guns and ingredients for illegal drugs.

Software supply chain
There are many open source frameworks for computer vision, navigation, etc. that can be used for carrying out attacks, and products often come with some built-in software for purposes such as flight stabilization. But not all powerful AI tools are widely distributed, or particularly easy to use currently. For example, large trained AI classification systems that reside within cloud computing stacks controlled by big companies (which are expensive to train), may be tempting for malicious actors to build from, potentially suggesting another point of control (discussed in Interventions and Appendix B).

Robot users
There are also registration requirements for some forms of robots such as drones in many countries, as well as requirements for pilot training, though we note that the space of robots that could cause physical harm goes beyond just drones. There are also no fly zones, imposed at a software level via manufacturers and governments, which are intended to prevent the use of consumer drones in certain areas, such as near airports, where the risk of unintentional or intentional collision between drones and passenger aircrafts looms large[1]. Indeed, at least one drone has already struck a passenger aircraft[2], suggesting a strong need for such no fly zones.

Governments
There is active discussion at the United Nations Convention on Certain Conventional Weapons of the value and complexity of banning or otherwise regulating lethal autonomous weapons systems[3]. Key states' opposition to a strong ban makes such an agreement unlikely in in the near-term, though the development of norms that could inform stronger governance is plausible[4]. Already in the United States, for example, there is an official Department of Defense directive that sets out policy for the development and use of autonomy in weapons[5]. Additionally, the U.S. Law of War Manual notes that humans are the primary bearers of responsibility for attacks in armed conflict[6]. The International Committee of the Red Cross has adopted a similar position, a stance that presumably implies some minimum necessary degree of human involvement in the use of force[7]. While such arms control discussions and norm development processes are critical, they are unlikely to stop motivated non-state actors from conducting attacks.

Physical defenses
In the physical sphere, there are many possible defenses against attacks via robots, though they are imperfect and unevenly distributed at present. Many are expensive and/or require human labor to deploy, and hence are only used to defend "hard targets"

1 Mouawad, 2015; Vincent, 2016
2 The Telegraph, 2016
3 Crootof, 2015
4 Crootof and Renz, 2017
5 DoD, 2012
6 DoD, 2015; Roff, 2016b
7 ICRC, 2017; Scharre, 2018

like safety-critical facilities and infrastructure (e.g. airports), the owners of which can afford to invest in such protection, as opposed to the much more widely distributed “soft targets” (such as highly populated areas). Physical defenses can include detection via radar, lidar, acoustic signature, or image recognition software[1]; interception through various means[2]; and passive defense through physical hardening or nets. The U.S. Department of Defense has recently launched a major program to defend against drones, and has tested lasers and nets with an eye towards defending against drones from the Islamic State in particular[3]. Given the potential for automation to allow attacks at scale, a particular challenge for defenders is finding effective methods of defense with an acceptable cost-exchange ratio[4]. As of yet, these defenses are incomplete and expensive, suggesting a likely near-term gap between the ease of attack and defense outside of heavily guarded facilities that are known targets (e.g. airports or military bases).

1 Aker and Kalkan, 2017

2 Yin, 2015; Scharre, 2015

3 Schmitt, 2017

4 Scharre, 2015

Payload control

An actor who wants to launch an aerial drone attack carrying a dangerous payload must source both the drone and the payload. Developed countries generally have long-lasting and reasonably effective systems to restrict access to potentially explosive materials, and are introducing systems to restrict access to acids (following high-profile acid attacks). More generally, state security and intelligence services uncover and foil a large number of attempted attacks, including those that involve attempts to procure dangerous materials. Increases in AI capabilities will likely help their work e.g. in analysing signal intelligence, or in characterising and tracking possible attackers.

## Political Security

Next, we discuss the political risks associated with malicious AI use. AI enables changes in the nature of communication between individuals, firms, and states, such that they are increasingly mediated by automated systems that produce and present content. Information technology is already affecting political institutions in myriad ways — e.g. the role of social media in elections, protests, and even foreign policy[5]. The increasing use of AI may make existing trends more extreme, and enable new kinds of political dynamics. Worryingly, the features of AI described earlier such as its scalability make it particularly well suited to undermining public discourse through the large-scale production of persuasive but false content, and strengthening the hand of authoritarian regimes. We consider several types of defenses, but as yet, as in the cases of Digital Security and Physical Security, the problem is unsolved.

5 Zeitzoff, 2017

## Context

There are multiple points of intersection between existing information technologies and the political sphere. Historically, politics and instability have had a symbiotic relationship with technological advances. Security needs have driven technological advances, and new technology has also changed the kinds of security threats that states and politicians face. Examples abound including the advent of the semaphore telegraph in Napoleonic France[1], to the advent of GPS and its use during the First Gulf War[2], to the use of social media during the Arab Spring[3]. Technological advances can change the balance of power between states, as well as the relationship between incumbent leaders and protesters seeking to challenge them. Modern militaries and intelligence agencies use today's information technologies for surveillance, as they did with previous generations of technologies such as telephones.

1 Schofield, 2013

2 Greenemeier, 2016

3 Aday et al., 2012

However, the effects of new technologies on these power relations are not straightforward. For example, social media technologies empower both incumbents and protesters: they allow military intelligences to monitor sentiment and attitudes, and to communicate more quickly; however, they also provide protesters in places such as Ukraine and Egypt, and rebel groups and revolutionary movements such as ISIS or Libyan rebels, the ability to get their message out to sympathetic supporters around the world[4] more quickly and easily. In addition, research suggests that social media may empower incumbent authoritarian regimes[5], as incumbent governments can manipulate the information that the public sees. Finally, some have argued that social media has further polarized political discourse, allowing users, particularly in the West, to self-select into their own echo chambers, while others have questioned this assumption[6]. Machine learning algorithms running on these platforms prioritize content that users are expected to like. Thus the dynamics we observe today are likely to only accelerate as these algorithms and AI become even more sophisticated.

4 Berger and Morgan, 2015; Jones and Mattiaci, 2017

5 Morozov, 2012; Rød and Weidmann 2015

6 Barberá et al., 2015

While they have evolved from previous technologies, information communication technologies are notable in some respects, such as the ease of information copying and transmission. Waltzmann writes, "The ability to influence is now effectively 'democratized,' since any individual or group can communicate and influence large numbers of others online"[7]. This "democratization" of influence is not necessarily favorable to democracy, however. It is very easy today to spread manipulative and false information, and existing approaches for detecting and stopping the spread of "fake news" fall short. Other structural aspects of modern technologies and the

7 Waltzmann, 2017

media industry also enable these trends. Marwick and Lewis (2017) note that the media's "dependence on social media, analytics and metrics, sensationalism, novelty over newsworthiness, and clickbait makes them vulnerable to such media manipulation." Others, such as Morozov (2012) and King, Pan, and Roberts (2017) argue that social media provides more tools for authorities to manipulate the news environment and control the message.

Finally, we note that the extent and nature of the use of information communication technologies to alter political dynamics varies across types of political regimes. In liberal democracies, it can be thought of as more of an emergent phenomenon, arising from a complex web of industry, government, and other actors, whereas in states like China, there is an explicit and deliberate effort to shape online and in-person political discussions, making use of increasingly sophisticated technologies to do so[1]. For instance, the Chinese government is exploring ways to leverage online and offline data to distill a "social credit score" for its citizens, and the generally more widespread use of censorship in China exemplifies the more explicit leveraging of technology for political purposes in some authoritarian states[2].

## How AI Changes the Political Security Landscape

AI will cause changes in the political security landscape, as the arms race between production and detection of misleading information evolves and states pursue innovative ways of leveraging AI to maintain their rule. It is not clear what the long-term implications of such malicious uses of AI will be, and these discrete instances of misuse only scratch the surface of the political implications of AI more broadly[3]. However, we hope that understanding the landscape of threats will encourage more vigorous prevention and mitigation measures.

Already, there are indications of how actors are using digital automation to shape political discourse. The widespread use of social media platforms with low barriers to entry makes it easier for AI systems to masquerade as people with political views. This has led to the widespread use of social media "bots" to spread political messages and cause dissent. At the moment, many such bots are controlled by humans who manage a large pack of bots[4], or use very simple forms of automation. However, these bot-based strategies (even when using relatively unsophisticated automation) are leveraged by national intelligence agencies and have demonstrated the ability to influence mainstream media coverage and political beliefs[5]. For instance, during both the Syrian Civil War[6] and the 2016 US election bots appeared to actively try to sway public opinion[7].

1 King, Pan, and Roberts, 2017

2 Botsman, 2017

3 It should be emphasised here again that we only consider in this report the direct malicious use of AI systems to undermine individual or collective security (see: Introduction). There are much larger systemic and political issues at stake with AI such as data aggregation and centralization, corporate/state control of the technology, legal and societal barriers to access and benefit, effects on employment, and issues relating to the economic and social distribution of risks and benefits, including aspects of equality. All of these are likely to have significant and complex effects on all aspects of political life, not just on political security. However, as outlined above, we set such system-wide risks outside the scope of this report.

4 Weedon et al., 2017

5 Woolley and Howard, 2017

6 Abokhodair et al., 2015

7 Guilbeault and Woolley, 2016

3 Adams, 2017; Serban et al., 2017

4 Seymour and Tully, 2016

5 Everett et al., 2016

6 Shao et al., 2017

7 Chung, Jamaludin, and Zisserman, 2017

8 Browne, 2017

Greater scale and sophistication of autonomous software actors in the political sphere is technically possible with existing AI techniques[3]. As previously discussed, progress in automated spear phishing has demonstrated that automatically generated text can be effective at fooling humans[4], and indeed very simple approaches can be convincing to humans, especially when the text pertains to certain topics such as entertainment[5]. It is unclear to what extent political bots succeed in shaping public opinion, especially as people become more aware of their existence, but there is evidence they contribute significantly to the propagation of fake news[6].

In addition to enabling individuals and groups to mislead the public about the degree of support for certain perspectives, AI creates new opportunities to enhance "fake news" (although, of course, propaganda does not require AI systems to be effective). AI systems may simplify the production of high-quality fake video footage of, for example, politicians saying appalling (fake) things[7]. Currently, the existence of high-quality recorded video or audio evidence is usually enough to settle a debate about what happened in a given dispute, and has been used to document war crimes in the Syrian Civil War[8]. At present, recording and authentication technology still has an edge over forgery technology. A video of a crime being committed can serve as highly compelling evidence even when provided by an otherwise untrustworthy source. In the future, however, AI-enabled high-quality forgeries may challenge the "seeing is believing" aspect of video and audio evidence. They might also make it easier for people to deny allegations against them, given the ease with which the purported evidence might have been produced. In addition to augmenting dissemination of misleading information, the writing and publication of fake news stories could be automated, as routine financial and sports reporting often are today. As production and dissemination of high-quality forgeries becomes increasingly low-cost, synthetic multimedia may constitute a large portion of the media and information ecosystem.

Even if bot users only succeed in decreasing trust in online environments, this will create a strategic advantage for political ideologies and groups that thrive in low-trust societies or feel opposed by traditional media channels. Authoritarian regimes in particular may benefit from an information landscape where objective truth becomes devalued and "truth" is whatever authorities claim it to be.

Moreover, automated natural language and multimedia production will allow AI systems to produce messages to be targeted at those most susceptible to them. This will be an extension of existing advertising practices. Public social media profiles are already reasonably predictive of personality details[1], and may be usable to predict psychological conditions like depression[2]. Sophisticated AI systems might allow groups to target precisely the right message at precisely the right time in order to maximize persuasive potential. Such a technology is sinister when applied to voting intention, but pernicious when applied to recruitment for terrorist acts, for example. Even without advanced techniques, "digital gerrymandering"[3] or other forms of advertising might shape elections in ways that undermine the democratic process.

The more entrenched position of authoritarian regimes offers additional mechanisms for control through AI that are unlikely to be as easily available in democracies[4]. AI systems enable fine-grained surveillance at a more efficient scale[5]. While existing systems are able to gather data on most citizens, efficiently using the data is too costly for many authoritarian regimes. AI systems both improve the ability to prioritise attention (for example, by using network analysis to identify current or potential leaders of subversive groups[6]) and also reduce the cost of monitoring individuals (for example, using systems that identify salient video clips and bring them to the attention of human agents). Furthermore, this can be a point of overlap between political and physical security, since robotic systems could also allow highly resourced groups to enforce a greater degree of compliance on unwilling populations.

The information ecosystem itself enables political manipulation and control by filtering content available to users. In authoritarian regimes, this could be done by the state or by private parties operating under rules and directions issued by the state. In democracies, the state may have limited legal authority to shape and influence information content but the same technical tools still exist; they simply reside in the hands of corporations. Even without resorting to outright censorship, media platforms could still manipulate public opinion by "de-ranking" or promoting certain content. For example, Alphabet Executive Chairman Eric Schmidt recently stated that Google would de-rank content produced by Russia Today and Sputnik[7]. In 2014, Facebook manipulated the newsfeeds of over half a million users in order to alter the emotional content of users' posts, albeit modestly[8]. While such tools could be used to help filter out malicious content or fake news, they also could be used by media platforms to manipulate public opinion[9].

Finally, the threats to digital and physical security that we have described in previous sections may also have worrying implications for political security. The hacking of the Clinton campaign in the 2016 presidential election is a recent example of how successful

1 Quercia et al., 2011; Kosinski et al., 2013

2 De Choudhury et al., 2013

3 De Choudhury et al., 2013

4 Roff, 2015c; Horowitz, 2016

5 Ng, 2015

6 Roff, 2015c

7 BBC, 2017

8 Goel, 2014; Kramer et al., 2014; Verma, 2014

9 Reuters, 2017; Tom Stocky, 2016; Griffith, 2017; Manjoo, 2017

cyberattacks can cause political disruption. The disruptive potential of physical attacks, such as assassinations and acts of terror, is even clearer. Such threats to digital and physical security might either undermine existing political institutions or allow them to justify a move toward more authoritarian policies.

## Points of Control and Existing Countermeasures

Several measures are already in development or deployed in this area, though none has yet definitively addressed the problems. We highlight a few relevant efforts here, and emphasize that these proposals are oriented towards the protection of healthy public discourse in democracies. Preventing more authoritarian governments from making full use of AI seems to be an even more daunting challenge.

Technical tools. Technical measures are in development for detecting forgeries[1] and social media bots[2]. Likewise, the use of certified authenticity of images and videos, e.g. the ability to prove that a video was broadcast live rather than synthesized offline[3] are valuable levers for ensuring that media is in fact produced by the relevant person or organization and is untampered in transit. Analogous measures have been developed for authentication of images (rather than videos) by Naveh and Tromer (2016).

Automated fake news detection is likewise the subject of ongoing research[4] as well as a competition, the Fake News Challenge[5], which can be expected to spur further innovation in this area. As yet, however, the detection of misleading news and images is an unsolved problem, and the pace of innovation in generating apparently authentic multimedia and text is rapid.

Pervasive use of security measures. Encryption is a generally useful measure for ensuring the security of information transmissions, and is actively used by many companies and other organizations, in part to prevent the sorts of risks discussed here. The use of citizens' data by intelligence agencies takes various forms and has been actively debated, especially in the wake of the Snowden revelations[6].

General interventions to improve discourse. There are various proposals to increase the quality of discourse in the public and private spheres, including longstanding ones such as better education and teaching of critical thinking skills, as well as newer ones ranging from tools for tracking political campaigning in social media (such as "Who Targets Me?"[7]) to policy proposals[8] to apps for encouraging constructive dialogue[9].

1 Kashyap et al., 2017; D'Avino et al., 2017

2 Varol et al, 2017

3 Rahman et al., 2017

4 Shu et al., 2017; Pérez-Rosas et al., 2017; Zubiaga et al., 2017

5 The Fake News Challenge is a competition aimed at fostering the development of AI tools to help human fact checkers combat fake news.

6 Clarke et al, 2013

7 "Who Targets Me?" is a software service that informs citizens on the extent with which they are being targeted by dark advertising campaigns.

8 Sunstein, 2017

9 Ixy, 2017

Media platforms. There have always been news sources of varying impartiality, and some online sources have better reputations than others, yet this has not entirely stopped the spread of fake news. Likewise, most people are aware of the existence of Ponzi schemes, scam emails, misleading sales tactics, etc. and yet victims are still found. Part of the reason that spam is less of a problem today than it otherwise could be is that the owners of key platforms such as email servers have deployed sophisticated spam filters. More generally, technology companies, social media websites, and and media organizations are critical points of control for stemming the tide of increasingly automated disinformation, censorship, and persuasion campaigns. Additionally, these organizations have unique datasets that will be useful for developing AI systems for detecting such threats, and through the ability to control access, they can pursue other strategies for preventing malicious uses of these platforms such as imposing strong barriers to entry (e.g. the use of one's offline identity) and limiting the rate at which accounts can disseminate information. Because these media platforms are for-profit corporations, public discourse, transparency, and potentially regulation will be important mechanisms for ensuring that their use of these powerful tools aligns with public interest[4].

4 Lapowsky, 2017

A development that occurred during the process of writing this report is illustrative. Late 2017 saw the rise of "deepfakes," the application of face-swapping algorithms to (among other applications) adult videos. While such videos first began appearing en masse in Reddit fora clearly labeled as being fictitious, the realism of some such deepfakes is an early sign of the potential decline of "seeing is believing" discussed above. After substantial media coverage of deepfakes, Reddit and other online websites, including adult content websites, began to crack down on the discussion and propagation of the technique. While these efforts have not been fully successful, they illustrate the critical role of technology platforms in governing information access, and it is likely that the deepfakes crackdown at least somewhat slowed the dissemination of the tool and its products, at least amongst less sophisticated actors.

# 04 Interventions

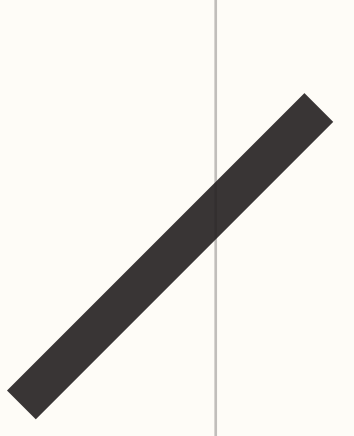

We identify a wide range of potential responses to the challenges raised above, as well as a large number of areas for further investigation. This section first makes several initial high-level recommendations for AI and ML researchers, policymakers, and others. We then suggest specific priority areas for further research, where investigation and analysis could develop and refine potential interventions to reduce risks posed by malicious use.

Due to the exploratory nature of this report, our primary aim is to draw attention to areas and potential interventions that we believe should be the subject of further investigation, rather than to make highly specific technical or policy proposals that may not be viable. The structure of this section, and the inclusion of Appendix B with additional exploratory material, is informed by this perspective.

## Recommendations

In this subsection we present four high-level recommendations, which are focused on strengthening the dialog between technical researchers, policymakers, and other stakeholders. In the following subsection, we will turn our attention to more concrete priority areas for technical work as well as associated research questions.

Our first pair of recommendations arise from the fact that the issues raised in this report combine technical and non-technical considerations, such as social, economic and military considerations. Concerns were raised at the workshop that the development of viable, appropriate responses to these issues may be hampered by two self-reinforcing factors: first, a lack of deep technical understanding on the part of policymakers, potentially leading to poorly-designed or ill-informed regulatory, legislative, or other policy responses; second, reluctance on the part of technical researchers to engage with these topics, out of concern that association with malicious use would tarnish the reputation of the field and perhaps lead to reduced funding or premature regulation. Our first two recommendations aim at preempting this dynamic.

Recommendation #1: Policymakers should collaborate closely with technical researchers to investigate, prevent, and mitigate potential malicious uses of AI. This must include policymakers taking seriously their responsibility to avoid implementing measures that will interfere with or impede research progress, unless those measures are likely to bring commensurate benefits. Close collaboration with technical experts also ensures that policy responses will be informed by the technical realities of the technologies at hand[1].

Recommendation #2: Researchers and engineers in artificial intelligence should take the dual-use nature of their work seriously, allowing misuse-related considerations to influence research priorities and norms, and proactively reaching out to relevant actors when harmful applications are foreseeable. Given that AI is a dual-use technology, we believe it is important that researchers consider it their responsibility to take whatever steps they can to help promote beneficial uses of the technology and prevent harmful uses. Example steps could include engaging with policymakers to provide expertise, and considering the potential applications of different research projects before deciding what to work on. (We recognize and appreciate the many AI researchers — including the technical experts who took part in the workshop and contributed to this report and other related initiatives — who are already doing outstanding work along these lines.[2])

1 Introductory resources for policymakers interested in this domain are increasingly becoming available, both generally about AI (Buchanan and Taylor, 2017), and specifically on AI and security (CNAS, 2017). As an example of policymaking in this domain that has surfaced several difficulties, the European Union's General Data Protection Regulation is a commonly-discussed example of a policy that is hard to interpret and apply in the context of current machine learning algorithms (Goodman and Flaxman, 2016).

2 The work of the Partnership on AI, the White House's 2016 series of workshops on AI, the 2017 "Beneficial AI" conference in Asilomar, and the AI Now conference series and organization are further examples where contributions from technical experts have been substantial and valuable.

We also make two recommendations laying out aims that we believe the broader AI community (including both technical and policy professionals) should work towards.

Recommendation #3: Best practices should be identified in research areas with more mature methods for addressing dual-use concerns, such as computer security, and imported where applicable to the case of AI. An example of a best practice that workshop participants considered clearly valuable to introduce into AI contexts is extensive use of "red teaming."[1] See Priority Research Area #1, below, for further details.

Recommendation #4: Actively seek to expand the range of stakeholders and domain experts involved in discussions of these challenges. This could include reaching out to sectors like civil society, national security experts, as-yet unengaged AI and cybersecurity researchers, businesses incorporating AI into their products, ethicists, the general public[2], and others, to ensure that relevant stakeholders are included and relevant experts consulted.

Because of the dual-use nature of AI, many of the malicious uses of AI outlined in this report have related legitimate uses. In some cases, the difference between legitimate and illegitimate uses of AI could be one of degree or ensuring appropriate safeguards against malicious use. For example, surveillance tools can be used to catch terrorists or oppress ordinary citizens. Information content filters could be used to bury fake news or manipulate public opinion. Governments and powerful private actors will have access to many of these AI tools and could use them for public good or harm. This is why a public dialogue on appropriate uses of AI technology is critical. The above four recommendations can help foster a cross-disciplinary dialogue among AI researchers, policymakers, and other relevant stakeholders to ensure that AI technology is used to benefit society.

1 In computer security, red teaming involves a "red team", composed of security experts and/or members of the host organization, deliberately planning and carrying out attacks against the systems and practices of the organization (with some limitations to prevent lasting damage), with an optional "blue team" responding to these attacks. These exercises explore what an actual attack might look like in order to better understand and, ultimately, improve the security of the organisation's systems and practices.

2 We expect adaptive defensive actions will be required of everyday citizens, if only in terms of maintaining awareness of threats and adopting best practices. It is important to acknowledge that different communities will have varying abilities to make such adaptations, depending for example on their technological literacy, which may pose challenges for implementing security policies. This is important not just for the communities less able to adapt to the new threats, but also for society more broadly as, for example, insecure systems may be compromised by attackers and repurposed to provide computing power and data for yet-more-capable attacks, while reducing the possibility that the attacks could be attributed, as they would then seem to originate from the compromised system.

## Priority Areas for Further Research

This section lays out specific topic areas that we recommend be investigated further. We aim here for brevity; more specific questions for investigation, along with additional context and commentary on many of the topics mentioned, may be found in Appendix B.

## Priority Research Area #1:
## Learning from and with the Cybersecurity Community

As AI-based systems become more widespread and capable, the potential impacts of cybersecurity incidents are growing commensurately. To summarize the considerations in the Digital Security section, AI is important to cybersecurity for three reasons. First, increased automation brings with it increased digital control of physical systems; consider, for example, how much more control a successful hacker could exercise over a modern car, compared with a typical car from 20 years ago[1]. Second, successful attacks on AI-based systems can also give the attacker access to the algorithms and/or trained models used by the system; consider, for example, theft of the datasets used for facial recognition on social networks, or the compromise of an algorithm used for analysing satellite imagery. Third, increasing use of AI in cyberattacks is likely to allow highly sophisticated attacks to be carried out at a much larger scale, which may reach victims that would otherwise not be suitable targets of previous waves of sophisticated attacks.

To respond to these increased dangers, cybersecurity must be a major and ongoing priority in efforts to prevent and mitigate harms from AI systems, and best practices from cybersecurity must be ported over wherever applicable to AI systems.

Some examples of cybersecurity-related sub-areas that we believe should be the subject of further research and analysis, then be implemented as appropriate (see Appendix B for more commentary on and questions about these sub-areas), include:

- Red teaming. Extensive use of red teaming to discover and fix potential security vulnerabilities and safety issues should be a priority of AI developers, especially in critical systems.
- Formal verification. To what extent, in what circumstances, and for what types of architectures can formal verification be used to prove key properties of AI systems? Can other approaches be developed to achieve similar goals by different means?[2]
- Responsible disclosure of AI vulnerabilities. Should AI-specific procedures be established to enable confidential reporting of vulnerabilities discovered in AI systems (including security vulnerabilities, potential adversarial inputs, and other types of exploits), as is already possible for security exploits in modern software systems?
- Forecasting security-relevant capabilities. Could "white-hat" efforts to predict how AI advances will enable more effective

1 For example, see the case of hackers first bringing a Jeep to a standstill on a busy highway, then later developing the ability to cause unintended acceleration and fully control the vehicle's steering (Greenberg, 2016).

2 DARPA's Assured Autonomy program (Neema, 2017) is one attempt at developing techniques to assure safety in systems that continue learning throughout their lifespans, which makes assurance or verification using traditional methods challenging. See also Katz et al., 2017; Selsam, Liang, and Dill, 2017; and Carlini et al., 2017.

cyberattacks, and more rigorous tracking of AI progress and proliferation[1] in general, allow for more effective preparations by defenders?

- Security tools. What tools (if any) should be developed and distributed to help make it standard to test for common security problems in AI systems, analogously to tools used by computer security professionals?

- Secure hardware. Could security features be incorporated into AI-specific hardware, for example to prevent copying, restrict access, facilitate activity audits, and similar? How technically and practically feasible is the design and adoption of hardware with properties like this?

1 Eckersley and Nasser et al., 2017

## Priority Research Area #2: Exploring Different Openness Models

Today, the prevailing norms in the machine learning research community strongly point towards openness. A large fraction of novel research is published online in papers that share anything from rough architectural outlines to algorithmic details to source code. This level of openness has clear benefits in terms of enabling researchers to build on each others' work, promoting collaboration, and allowing theoretical progress to be incorporated into a broad array of applications.

However, the potential misuses of AI technology surveyed in the Scenarios and Security Domains sections suggest a downside to openly sharing all new capabilities and algorithms by default: it increases the power of tools available to malicious actors. This raises an important research question: might it be appropriate to abstain from or merely delay publishing of some findings related to AI for security reasons? There is precedent for this in fields such as computer security, where exploits that could affect important systems are not publicly disclosed until the developers have had an opportunity to fix the vulnerability. To the extent that research results are withheld today in AI, it is usually for reasons related to intellectual property (e.g. in order to avoid a future result being "scooped"). In light of risks laid out elsewhere in this report, there may also be arguments based on public interest for additional caution in at least some cases.

While the proposals below consider decreasing openness in certain situations, we stress that there are clear and well-recognized reasons to favor openness in research communities. We believe that policies leading to decreased openness, while potentially appropriate in certain instances, should be sensitive to these benefits.[1] Rather than propose a specific solution, our aim is to foster discussion of whether and when considerations against open sharing might outweigh considerations in favor and what mechanisms might enable this.

Some potential mechanisms and models that could be subject to further investigation and analysis (see Appendix B for more commentary on and questions about for these sub-areas) include:

- Pre-publication risk assessment in technical areas of special concern. Should some types of AI research results, such as work specifically related to digital security or adversarial machine learning, be subject to some kind of risk assessment to determine what level of openness is appropriate[2]? This is the norm for research in other areas, such as biotechnology and computer security. Or would such measures be premature today, before AI systems are more widely used in critical systems and we have better knowledge of which technical research is most security-relevant? If such measures are considered be premature, under what conditions would they be appropriate?

- Central access licensing models. Could emerging "central access" commercial structures — in which customers use services like sentiment analysis or image recognition made available by a central provider without having access to the technical details of the system — provide a template for a security-focused sharing model that allows widespread use of a given capability while reducing the possibility of malicious use? How might such a model remain viable over time as advances in processing power, data storage and availability, and embedded expertise allow a larger set of actors to use AI tools?

- Sharing regimes that favor safety and security. Could arrangements be made under which some types of research results are selectively shared among a predetermined set of people and organizations that meet certain criteria, such as effective information security and adherence to appropriate ethical norms? For example, certain forms of offensive cybersecurity research that leverage AI might be shared between trusted organizations for vulnerability discovery purposes, but would be harmful if more widely distributed.

- Other norms and institutions that have been applied to dual-use technologies. What can be learned from other models, methodologies, considerations, and cautions that have arisen when tackling similar issues raised by other dual-use technologies?

1 Accordingly, concerns about misuse should not be used as an excuse to reduce openness to a greater extent than is required, for instance, when the real motivation is about corporate competitiveness. We believe that, to the extent that practices around openness are rethought, this should be done transparently, and that when new approaches are incorporated into AI research and publication processes from other domains (e.g. responsible disclosure), those doing so should state their reasons publicly so that a range of stakeholders can evaluate these claims. The debate in the biosecurity community about the appropriate level of disclosure on gain-of-function research (in which organisms are made more dangerous in order to understand certain threats better) provides a model of the kind of discussion we see as healthy and necessary.

2 see e.g. NDSS, 2018

## Priority Research Area #3: Promoting a Culture of Responsibility

AI researchers and the organizations that employ them are in a unique position to shape the security landscape of the AI-enabled world. Many in the community already take their social responsibility quite seriously, and encourage others to do the same. This should be continued and further developed, with greater leveraging of insights from the experiences of other technical fields, and with greater attentiveness to malicious use risks in particular. Throughout training, recruitment, research, and development, individuals, and institutions should be mindful of the risks of malicious uses of AI capabilities.

Some initial areas to explore for concrete initiatives aimed at fostering a culture of responsibility include[1]:

- Education. What formal and informal methods for educating scientists and engineers about the ethical and socially responsible use of their technology are most effective? How could this training be best incorporated into the education of AI researchers?
- Ethical statements and standards. What role should ethical statements and standards play in AI research?[2] How and by whom should they be implemented and enforced? What are the domain-specific ethical questions in the areas of digital, physical, and security that need to be resolved in order to distinguish between benign and malicious uses of AI?
- Whistleblowing measures. What is the track record of whistleblowing protections in other domains, and how (if at all) might they be used for preventing AI-related misuse risks?
- Nuanced narratives. More generally, are there succinct and compelling narratives of AI research and its impacts that can balance optimism about the vast potential of this technology with a level-headed recognition of the risks it poses? Examples of existing narratives include the “robot apocalypse” trope and the countervailing “automation boon” trope, both of which have obvious shortcomings. Might a narrative like “dual-use” (proposed above) be more productive?

1 See Appendix B for more commentary on and questions about these sub-areas.

2 Two examples of proposed standards for AI are the IEEE Global Initiative for Ethical Considerations in Artificial Intelligence and Autonomous Systems (IEEE Standards Association, 2017) and the development of the Asilomar AI Principles (Future of Life Institute, 2017).

## Priority Research Area #4:
## Developing Technological and Policy Solutions

In addition to creating new security challenges and threats progress in AI also makes possible new types of responses and defenses. These technological solutions must be accompanied and supported by well-designed policy responses. In addition to the proposals mentioned in the previous sections, what other potential approaches — both institutional and technological — could help to prevent and mitigate potential misuse of AI technologies?

Some initial suggested areas for further investigation include[1]:

- Privacy protection. What role can technical measures play in protecting privacy from bad actors in a world of AI? What role must be played by institutions, whether by corporations, the state, or others?
- Coordinated use of AI for public-good security. Can AI-based defensive security measures be distributed widely to nudge the offense-defense balance in the direction of defense?[2] Via what institutions or mechanisms can these technologies be promoted and shared?
- Monitoring of AI-relevant resources. Under what circumstances, and for which resources, might it be feasible and appropriate to monitor inputs to AI technologies such as hardware, talent, code, and data?
- Other legislative and regulatory responses. What other potential interventions by policymakers would be productive in this space (e.g. adjusting legal definitions of hacking to account for the case of adversarial examples and data poisoning attacks)?

For all of the above, it will be necessary to incentivize individuals and organizations with the relevant expertise to pursue these investigations. An initial step, pursued by this report, is to raise awareness of the issues and their importance, and to lay out an initial research agenda. Further steps will require commitment from individuals and organizations with relevant expertise and a proven track record. Additional monetary resources, both public and private, would also help to seed interest and recruit attention in relevant research communities.

1 See Appendix B for more commentary on and questions about these sub-areas.

2 For example, could AI systems be used to refactor existing code bases or new software to adhere more closely to principle of least authority (Miller, 2006) or other security best practices?

# 05 Strategic Analysis

When considered together, how will the security-relevant characteristics of AI and the various intervention measures surveyed above (if implemented) combine to shape the future of security? Any confident long-term prediction is impossible to make, as significant uncertainties remain regarding the progress of various technologies, the strategies adopted by malicious actors, and the steps that should and will be taken by key stakeholders. Nonetheless, we aim to elucidate some crucial considerations for giving a more confident answer, and make several hypotheses about the medium-term equilibrium of AI attack and defense. By medium-term, we mean the time period (5+ years from now) after which malicious applications of AI are widely used and defended against, but before AI has yet progressed sufficiently to fully obviate the need for human input in either attack or defense.

Even a seemingly stable and predictable medium-term equilibrium resulting from foreseeable AI developments might be short-lived, since both technological and policy factors will progress beyond what can currently be foreseen. New developments, including technological developments unrelated to AI, may ultimately be more impactful than the capabilities considered in this report. Nevertheless, we hope that the analysis below sheds some light on key factors to watch and influence in the years to come.

## Factors Affecting the Equilibrium of AI and Security

### Attacker Access to Capabilities

Current trends emphasize widespread open access to cutting-edge research and development achievements. If these trends continue for the next 5 years, we expect the ability of attackers to cause harm with digital and robotic systems significantly increase. This follows directly from the dual-use nature, efficiency, scalability, and ease of diffusing AI technologies discussed previously.

However, we expect the dual-use nature of the technology will become increasingly apparent to developers and regulators, and that limitations on access to or malicious use of powerful AI technologies will be increasingly imposed. However, significant uncertainty remains about the effectiveness of attempting to restrict or monitor access through any particular intervention. Preemptive design efforts and the use of novel organizational and technological measures within international policing will all help, and are likely to emerge at various stages, in response (hopefully) to reports such as these, or otherwise in the aftermath of a significant attack or scandal. Efforts to prevent malicious uses solely through limiting AI code proliferation are unlikely to succeed fully, both due to less-than-perfect compliance and because sufficiently motivated and well resourced actors can use espionage to obtain such code. However, the risk from less capable actors using AI can likely be reduced through a combination of interventions aimed at making systems more secure, responsibly disclosing developments that could be misused, and increasing threat awareness among policymakers.

### Existence of AI-Enabled Defenses

The same characteristics of AI that enable large-scale and low-cost attacks also allow for more scalable defenses. Specific instances of AI-enabled defenses have been discussed in earlier sections, such as spam filters and malware detection, and we expect many others will be developed in the coming years. For example, in the context of physical security, the use of drones whose sole purpose is to quickly and non-violently "catch" and bring to the ground other drones might be invented and widely deployed, but they might also turn out to be prohibitively expensive, as might other foreseeable defenses. Thus, both the pace of technical innovation and the cost of such defenses should be considered in a fuller assessment.

One general category of AI-enabled defenses worth considering in an overall assessment is the use of AI in criminal investigations and counterterrorism. AI is already beginning to see wider adoption for a wide range of law enforcement purposes, such as facial recognition by surveillance cameras and social network analysis. We have hardly seen the end of such advancements, and further developments in the underlying technologies and their widespread use seem likely given the interest of actors from corporations to governments in preventing criminal acts. Additionally, interceding attacks in their early stage through rapid detection and response may turn out to be cheaper than for example widely deploying physical defenses against drones. Thus, the growing ability of states to detect and stop criminal acts, in part by leveraging AI, is a key variable in the medium-term. However, such advances will not help prevent authoritarian abuses of AI.

## Distribution and Generality of Defenses

Some defensive measures discussed in Interventions and Appendix B can be taken by single, internally coordinated actors, such as research labs and tech startups, and are likely to happen as soon as they become technically feasible and cost-effective. These measures could then be used by the organizations that have the most to lose from attacks such as governments and major corporations. This means that the most massive category of harm, such as attack on WMD facilities, is also the least likely, though the level of risk will depend on the relative rates at which attacks and defenses are developed. Responsible disclosure of novel vulnerabilities, pre-publication risk assessment, and a strong ethical culture in the AI community more generally will be vital in such a world.

This, however, leaves out the strategic situation for the majority of potential victims: technologically conservative corporations, under-resourced states, SMEs, and individuals. For these potential victims, defensive measures need to be baked into widespread technology, which may require coordinated regulatory efforts, or offered at low prices. The latter is most likely to come either from tech giants (as in the case of spam filters), which will increase lock-in and concentration of data and power, or from non-profit organizations who develop and distribute such defensive measures freely or cheaply (e.g. Mozilla's Firefox web browser).

This dynamic of defense through reliance on fortified software platforms is likely to be affected by the generality of defensive measures: if each attack requires a tailored defense, and has an associated higher time lag and skill investment, it is more likely that those developing such defensive measures will need financial backing, from corporations, investors, philanthropists, or governments. In the case of governments, international competition may hinder the development and release of defensive measures, as is generally the case in cyber-security, though see the release of CyberChef[1] and Assemblyline[2] as counterexamples. For political security, similar considerations regarding generality apply: a general solution to authenticable multimedia production and forgery detection would be more useful than tailored individual solutions for photographs, videos, or audio, or narrower subsets of those media types.

1 GCHQ, 2016

2 CSE, 2017

Misaligned incentives can also lead to a failure to employ available defensive measures. For example, better cybersecurity defenses could raise the bar for data breaches or the creation of IoT device botnets. However, the individuals affected by these failures, such as the individuals whose personal data is released or victims of DDOS attacks using botnets, are not typically in a position to improve defenses directly. Thus, other approaches including regulation may be needed to adjust these incentives or otherwise address these externalities[3].

3 Moore and Anderson, 2012

## Overall Assessment

The range of plausible outcomes is extremely diverse, even without considering the outcomes that are less likely, but still possible. Across all plausible outcomes, we anticipate that attempts to use AI maliciously will increase alongside the increase in the use of AI across society more generally. This is not a trend that is particular to AI; we anticipate increased malicious use of AI just as criminals, terrorists and authoritarian regimes use electricity, software, and computer networks: at some point in the technology adoption cycle, it becomes easier to make use of such general purpose technologies than to avoid them.

On the optimistic side, several trends look positive for defense. There is much low hanging fruit to be picked in securing AI systems themselves, and securing people and systems from AI-enabled attacks. Examples include responsible vulnerability disclosure for machine learning in cases where the affected ML technology is being used in critical systems, and greater efforts to leverage AI expertise in the discovery of vulnerabilities by software companies internally before they are discovered by adversaries. There are substantial academic incentives to tackle the hardest research problems, such as developing methods to address adversarial examples and providing provable guarantees for system properties and behaviors. There are, at least in some parts of the world, political incentives for developing processes and regulations that reduce threat levels and increase stability, e.g. through consumer protection and standardization. Finally, there are incentives for tech giants to collaborate on ensuring at least a minimal level of security for their users. Where solutions are visible, require limited or pre-existing coordination, and align with existing incentive structures, defenses are likely to prevail.

On the pessimistic side, not all of the threats identified have solutions with these characteristics. It is likely to prove much harder to secure humans from manipulation attacks than it will be to secure digital and cyber-physical systems from cyber attacks, and in some scenarios, all three attack vectors may be combined. In the absence of significant effort, attribution of attacks and penalization of attackers is likely to be difficult, which could lead to an ongoing state of low- to medium-level attacks, eroded trust within societies, between societies and their governments, and between governments. Whichever vectors of attack prove hardest to defend against will be the ones most likely to be weaponized by governments, and the proliferation of such offensive capability is likely to be broad. Since the number of possible attack surfaces is vast, and the cutting edge of capability is likely to be ever progressing, any equilibrium obtained between rival states or between criminals and security forces in a particular domain is likely to be short-lived as technology and policies evolve.

Tech giants and media giants may continue to become technological safe havens of the masses, as their access to relevant real-time data at massive scale, and their ownership of products and communication channels (along with the underlying technical infrastructure), place them in a highly privileged position to offer tailored protection to their customers. Other corporate giants that offer digitally-enhanced products and services (automotive, medical, defense, and increasingly many other sectors) will likely be under pressure to follow suit. This would represent a continuation of existing trends in which people very regularly interact with and use the platforms provided by tech and media giants, and interact less frequently with small businesses and governments.

Nations will be under pressure to protect their citizens and their own political stability in the face of malicious uses of AI[1]. This could occur through direct control of digital and communication infrastructure, through meaningful and constructive collaboration between the government and the private entities controlling such infrastructure, or through informed and enforceable regulation coupled with well-designed financial incentives and liability structures. Some countries have a clear head start in establishing the control mechanisms that will enable them to provide security for their citizens[2].

For some of the more challenging coordination and interdisciplinary problems, new leadership will be required to rise above local incentives and provide systemic vision. This will not be the first time humanity has risen to meet such a challenge: the NATO conference at Garmisch in 1968 created consensus around the growing risks from software systems, and sketched out technical and procedural solutions to address over-run, over-budget, hard-to-maintain and bug-ridden critical infrastructure software, resulting in many practices which are now mainstream in software engineering[3]; the NIH conference at Asilomar in 1975 highlighted the emerging risks from recombinant DNA research, promoted a moratorium on certain types of experiments, and initiated research into novel streams of biological containment, alongside a regulatory framework such research could feed into[4]. Individuals at the forefront of research played key roles in both of these cases, including Edsger Dijkstra in the former[5] and Paul Berg in the latter[6].

There remain many disagreements between the co-authors of this report, let alone amongst the various expert communities out in the world. Many of these disagreements will not be resolved until we get more data as the various threats and responses unfold, but this uncertainty and expert disagreement should not paralyse us from taking precautionary action today. Our recommendations, stated above, can and should be acted on today: analyzing and (where appropriate) experimenting with novel openness models, learning from the experience of other scientific disciplines, beginning multi-stakeholder dialogues on the risks in particular domains, and accelerating beneficial research on myriad promising defenses.

1 Chessen, 2017b

2 For example, France's campaign laws prohibited Macron's opponent from further campaigning once Macron's emails had been hacked. This prevented the campaign from capitalizing on the leaks associated with the hack, and ended up with the hack playing a much more muted role in the French election than the Clinton hack played in the US election.

3 Naur and Randell, 1969

4 Krimsky, 1982; Wright, 1994

5 Dijkstra, 1968

6 Berg et al., 1974

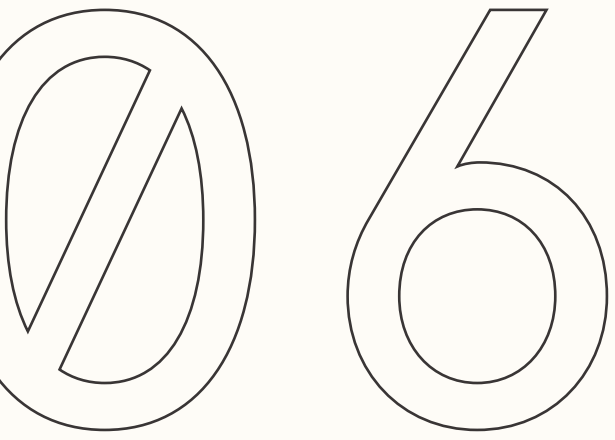

# Conclusion

While many uncertainties remain, it is clear that AI will figure prominently in the security landscape of the future, that opportunities for malicious use abound, and that more can and should be done.

Artificial intelligence, digital security, physical security, and political security are deeply connected and will likely become more so. In the cyber domain, even at current capability levels, AI can be used to augment attacks on and defenses of cyberinfrastructure, and its introduction into society changes the attack surface that hackers can target, as demonstrated by the examples of automated spear phishing and malware detection tools discussed above. As AI systems increase in capability, they will first reach and then exceed human capabilities in many narrow domains, as we have already seen with games like backgammon, chess, Jeopardy!, Dota 2, and Go and are now seeing with important human tasks like investing in the stock market or driving cars. Preparing for the potential malicious uses of AI associated with this transition is an urgent task.

As AI systems extend further into domains commonly believed to be uniquely human (like social interaction), we will see more sophisticated social engineering attacks drawing on these capabilities. These are very difficult to defend against, as even cybersecurity experts can fall prey to targeted spear phishing emails. This may cause an explosion of network penetrations, personal data theft, and an epidemic of intelligent computer viruses. One of our best hopes to defend against automated hacking is also via AI, through automation of our cyber-defense systems, and indeed companies are increasingly pursuing this strategy. But AI-based defense is not a panacea, especially when we look beyond the digital domain. More work should also be done in understanding the right balance of openness in AI, developing improved technical measures for formally verifying the robustness of systems, and ensuring that policy frameworks developed in a less AI-infused world adapt to the new world we are creating.

Looking to the longer term, much has been published about problems which might arise accidentally as a result of highly sophisticated AI systems capable of operating at high levels across a very wide range of environments[1], though AI capabilities fall short of this today. Given that intelligence systems can be deployed for a range of goals[2], highly capable systems that require little expertise to develop or deploy may eventually be given new, dangerous goals by hacking them or developing them de novo: that is, we may see powerful AI systems with a "just add your own goals" property. Depending on whose bidding such systems are doing, such advanced AIs may inflict unprecedented types and scales of damage in certain domains, requiring preparedness to begin today before these more potent misuse potentials are realizable. Researchers and policymakers should learn from other domains with longer experience in preventing and mitigating malicious use to develop tools, policies, and norms appropriate to AI applications.

Though the specific risks of malicious use across the digital, physical, and political domains are myriad, we believe that understanding the commonalities across this landscape, including the role of AI in enabling larger-scale and more numerous attacks, is helpful in illuminating the world ahead and informing better prevention and mitigation efforts. We urge readers to consider ways in which they might be able to advance the collective understanding of the AI-security nexus, and to join the dialogue about ensuring that the rapid development of AI proceeds not just safely and fairly but also securely.

1 Bostrom, 2014; Amodei and Olah et al., 2016

2 Bostrom, 2014, p. 107

## Acknowledgements

We are extremely grateful to the many researchers and practitioners who have provided useful comments on earlier versions of this document, and who engaged us in helpful conversations about related topics. Given the number of coauthors and related conversations, we will surely forget some people, but among others, we thank Ian Goodfellow, Ross Anderson, Nicholas Papernot, Martín Abadi, Tim Hwang, Laura Pomarius, Tanya Singh Kasewa, Smitha Milli, Itzik Kotler, Andrew Trask, Siddharth Garg, Martina Kunz, Jade Leung, Katherine Fletcher, Jan Leike, Toby Ord, Nick Bostrom, Owen Cotton-Barratt, Eric Drexler, Julius Weitzdorfer, Emma Bates, and Subbarao Kambhampati. Any remaining errors are the responsibility of the authors. This work was supported in part by a grant from the Future of Life Institute.

# Appendix A: Workshop Details

## Summary

On February 19 and 20, 2017, Miles Brundage of the Future of Humanity Institute (FHI) and Shahar Avin of the Centre for the Study of Existential Risk (CSER) co-chaired a workshop entitled "Bad Actor Risks in Artificial Intelligence" in Oxford, United Kingdom. The workshop was co-organized by FHI, CSER, and the Leverhulme Centre for the Future of Intelligence (CFI). Participants came from a wide variety of institutional and disciplinary backgrounds, and analyzed a variety of risks related to AI misuse. The workshop was held under Chatham House rules.

## Event Structure

On February 19, the event began with background presentations on cybersecurity, AI, and robotics from relevant experts in these fields. A particular focus of the presentations was on highlighting underexplored risks. The afternoon featured two sets of breakout sessions: participants first discussed security domains and scenarios, and then discussed possible defenses.

On February 20, a subset of the participants from the first day of the workshop met to discuss next steps and the prioritization of possible prevention and mitigation measures. The group present agreed upon the need for a research agenda to be produced, and voted on which measures seemed useful and tractable, in order to focus the subsequent report writing process.

## Report Writing Process

This document is based in large part on notes from the discussions at the workshop, as well as prior and subsequent research by the authors on the topic. Brundage and Avin et al. wrote a draft of the report and circulated it among all of the attendees at the workshop as well as additional domain experts. We are grateful to all of the workshop participants for their invaluable contributions, even if we were not able to capture all of their perspectives.

## List of Workshop Participants

Dario Amodei, OpenAI
Ross Anderson, University of Cambridge
Stuart Armstrong, Future of Humanity Institute
Amanda Askell, Centre for Effective Altruism
Shahar Avin, Centre for the Study of Existential Risk
Miles Brundage, Future of Humanity Institute
Joanna Bryson, University of Bath/Princeton University Center for Information Technology Policy
Jack Clark, OpenAI
Guy Collyer, Organization for Global Biorisk Reduction
Owen Cotton-Barratt, Future of Humanity Institute
Rebecca Crootof, Yale Law School
Allan Dafoe, Yale University
Eric Drexler, Future of Humanity Institute
Peter Eckersley, Electronic Frontier Foundation
Ben Garfinkel, Future of Humanity Institute
Carrick Flynn, Future of Humanity Institute
Ulrike Franke, University of Oxford
Dylan Hadfield-Menell, UC Berkeley and Center for Human-compatible AI
Richard Harknett, University of Oxford/University of Cincinnati
Katja Hofmann, Microsoft Research
Tim Hwang, Google

Eva Ignatuschtschenko, University of Oxford
Victoria Krakovna, DeepMind/Future of Life Institute
Ben Laurie, DeepMind
Jan Leike, DeepMind/Future of Humanity Institute
Seán Ó hÉigeartaigh, Centre for the Study of Existential Risk
Toby Ord, Future of Humanity Institute
Michael Page, Centre for Effective Altruism
Heather Roff, University of Oxford/Arizona State University /New America Foundation
Paul Scharre, Center for a New American Security
Eden Shochat, Aleph VC
Jaan Tallinn, Centre for the Study of Existential Risk
Helen Toner, Open Philanthropy Project
Andrew Trask, University of Oxford
Roman Yampolskiy, University of Louisville
Yueh-Hsuan Weng, Tohoku University

# Appendix B: Questions for Further Research

This appendix gives additional commentary on topics related to the Recommendations and Priority Research Areas described in the Interventions section of the main report, along with some initial questions and directions for investigation on each topic. In each case, we flag which one or more of the three high-level threat factors (introduced in General Implications for the Threat Landscape) the research area aims to address. We include this content as a jumping-off point for researchers interested in making progress in these areas; the below is not intended to be exhaustive or conclusive.

## Dual Use Analogies and Case Studies

One possible area of theory, practice, and history to be explored for insights is the set of technologies with prominent concerns around dual use - technologies that can be used for both peaceful and military aims (or, more generally, to both beneficial and harmful ends). Examples include chemicals potentially useful for chemical weapons or explosives, biological engineering potentially useful for biological weapons, cryptography, and nuclear technologies. Allen and Chan[1] explored several of these case studies and their potential insights for AI dual-use policymaking. In these cases, there is a rich tapestry of soft norms (e.g. pre-publication review) and hard laws (e.g. export controls) developed over many years to ensure positive outcomes[2].

When consulting the history of governing dual use technologies, we should learn both constructive solutions from past successes, and precautionary lessons about poor regulation that should be avoided. A relevant example of the latter is the difficulties of regulating cryptographic algorithms and network security tools through export control measures such as the Wassenaar Arrangement[3]. The similarities between AI and cryptography, in terms of running on general-purpose hardware, in terms of being immaterial objects (algorithms), in terms of having a very wide range of legitimate applications, and in their ability to protect as well as harm, suggest that the default control measures for AI might be similar those that have been historically applied to cryptography. This may well be a path we should avoid, or at least take very cautiously.

The apparent dual-use nature of AI technologies raises the following questions:

- What is the most appropriate level of analysis and governance of dual-use characteristics of AI technologies (e.g. the field as a whole, individual algorithms, hardware, software, data)?
- What norms from other dual-use domains are applicable to AI?
- What unique challenges, if any, does AI pose as a dual-use technology?
- Are there exemplary cases in which dual-use concerns were effectively addressed?
- What lessons can be learned from challenges and failures in applying control measures to dual-use technologies?

1 Allen and Chan, 2017

2 Tucker, ed. 2012; Harris, ed. 2016

3 Shehadeh, 1999

## Red Teaming

A common tool in cybersecurity and military practice is red teaming - a "red team" composed of security experts and/or members of the organization deliberately plans and carries out attacks against the systems and practices of the organization (with some limitations to prevent lasting damage), with an optional "blue team" responding to these attacks. These exercises explore what an actual attack might look like in order to ultimately better understand and improve the security of the organization's systems and practices. Two subsets of the AI security domain seem particularly amenable to such exercises: AI-enabled cyber offense and defense, and adversarial machine learning. While we highlight these subsets because they seem especially relevant to security, red teaming of AI technologies more broadly seems generally beneficial. In addition to this report and the associated workshop, another recent effort aimed at this goal was also conducted by the Origins Project earlier this year[1].

In the case of cyber attacks, many of the concerns discussed earlier in this document, and elsewhere in the literature, are hypothetical. Conducting deliberate red team exercises might be useful in the AI/cybersecurity domain, analogous to the DARPA Cyber Grand Challenge but across a wider range of attacks (e.g. including social engineering, and vulnerability exploitation beyond memory attacks), in order to better understand the skill levels required to carry out certain attacks and defenses, and how well they work in practice.

Likewise, in the case of adversarial machine learning, while there are many theoretical papers showing the vulnerabilities of machine learning systems to attack, the systematic and ongoing stress-testing of real-world AI systems has only just begun[2]. Efforts like the CleverHans library of benchmarks and models are a step in this direction[3], creating the foundation for a distributed open source red teaming effort, as is the NIPS 2017 Adversarial Attacks and Defenses competition[4], which is more analogous to the DARPA Cyber Grand Challenge.

There are several open questions regarding the use of "red team" strategies for mitigating malicious uses of AI:

- What lessons can be learned from the history to date of "red team" exercises?
- Is it possible to detect most serious vulnerabilities through "red team" exercises, or is the surface area for attack too broad?

1 Bass, 2017

2 though see e.g. Anderson et al., 2017

3 Papernot et al., 2016b

4 Knight, 2017

- Who should be responsible for conducting such exercises, and how could they be incentivised to do so?
- What sorts of skills are required to undermine AI systems, and what is the distribution of those skills? To what extent do these skills overlap with the skills required to develop and deploy AI systems, and how should these findings inform the threat model used in red teaming exercises (and other AI security analysis)?
- Are there mechanisms to promote the uptake of lessons from "red team" exercises?
- Are there mechanisms to share lessons from "red team" exercises with other organizations that may be susceptible to similar attacks? How to avoid disclosure of attack methods to bad actors?
- What are the challenges and opportunities of extending "red teaming" (or related practices like tabletop exercises) to AI issues in the physical and political domains? What can be learned for the physical domain from physical penetration testing exercises?

## Formal Verification

Formal verification of software systems has been studied for decades[1]. In recent years, it has been shown that even some very complex systems are amenable to formal proofs that they will operate as intended, including the CompCert compiler and the seL4 microkernel[2]. An open question is whether AI systems, or elements thereof, are amenable to formal verification. At the workshop there was substantial skepticism about the prospects for formal AI verification, given the complexity of some modern AI systems, but further analysis about the challenges is required, and research on the topic continues apace[3]. In particular, we might be interested in the following properties being verified for a given system:

- that its internal processes in fact attain the goals specified for the system (though noting the existence of the specification problem, i.e. that desired properties of AI systems are often difficult to specify in advance, and therefore difficult to verify),
- that its goals will be remain constant in the face of adversaries attempts to change them,

1 Turing, 1949; Baier and Katoen, 2008

2 Fisher, 2014

3 e.g. Selsam et al., 2017; Neema, 2017

- that its ability to be deceived with adversarial inputs is bounded to some extent.

### Verifying Hardware

Given the increasing complexity of AI systems, and in some domains limited theoretical foundations for their operation, it may be prohibitively expensive, or even practically or theoretically impossible, to provide an end-to-end verification framework for them. However, it may be feasible to use formal methods to improve the security of components of these systems. Hardware seems particularly amenable to verification, as formal methods have been widely adopted in the hardware industry for decades[1].

### Verifying Security

Additionally, in recent years formal verification has been applied to security protocols to provide robust guarantees of safety against certain types of attacks. The JavaScript prover CryptoVerif[2] is an example of a developer-focused tool that allows programmers to apply formal methods to their code to check correctness in the development process. It should be noted that much of this work is still largely theoretical and adoption in the real world has so far been limited[3].

### Verifying AI Functionality

The notion of being able to prove that a system behaves as intended is an attractive one for artificial intelligence. However, formal methods are difficult to scale up to arbitrary complex systems due to the state space explosion problem. Nonetheless, verification of some aspects of AI systems, such as image classifiers, is still feasible even verification of the behavior of the whole system is prohibitively complex. For example, work on verification of deep neural networks provided a method to check for the existence of adversarial examples in regions of the input space[4].

## Responsible "AI 0-Day" Disclosure

As discussed above, despite the successes of contemporary machine learning algorithms, it has been shown time and again that ML algorithms also have vulnerabilities. These include ML-specific vulnerabilities, such as inducing misclassification via adversarial

1 Harrison, 2010; Wahby, R. 2016

2 Blanchet, 2017

3 though there are some instances of real world use - see e.g. Beurdouche et al., 2017

4 Katz et al., 2017

examples[1] or via poisoning the training data[2]; see Barreno et al. (2010) for a survey. ML algorithms also remain open to traditional vulnerabilities, such as memory overflow (Stevens et al., 2016). There is currently a great deal of interest among cyber-security researchers in understanding the security of ML systems, though at present there seem to be more questions than answers.

In the cybersecurity community, “0-days” are software vulnerabilities that have not been made publicly known (and thus defenders have zero days to prepare for an attack making use of them). It is common practice to disclose these vulnerabilities to affected parties before publishing widely about them, in order to provide an opportunity for a patch to be developed.

Should there be a norm in the AI community for how to disclose such vulnerabilities responsibly to affected parties (such as those who developed the algorithms, or are using them for commercial applications)? This broad question gives rise to additional questions for further research:

- As AI technologies become increasingly integrated into products and platforms, will the existing security norm around responsible disclosure extend to AI technologies and communities?
- Should AI systems (both existing and future) be presumed vulnerable until proven secure, to an extent that disclosing new vulnerabilities privately is unnecessary?
- In what safety-critical contexts are AI systems currently being used?
- Which empirical findings in AI would be useful in informing an appropriate disclosure policy (analogous to the way that historical trends in 0-day discoveries and exploitation rates are discussed in cybersecurity analyses[3])?
- If such a norm were appropriate in broad terms, who should be notified in case a vulnerability is found, how much notice should be given before publication, and what mechanisms should institutions create to ensure a recommendation is processed and potentially acted upon?
- What is the equivalent of “patching” for AI systems, and how should trade-offs (e.g. between resource demands, accuracy and robustness to noise) and prioritization amongst the variety of possible defense measures be weighed in a world of rapidly changing attacks and defenses?

1 Szegedy et al., 2013; Papernot et al., 2016; Evtimov et al., 2017; Carlini et al., 2016

2 Rubinstein et al., 2009; Šrndic and Laskov, 2014

3 e.g. Ablon and Bogart, 2017

## AI-Specific Exploit Bounties

To complement the norm of responsible disclosure of vulnerabilities (discussed above), which relies on social incentives and goodwill, some software vendors offer financial incentives (cash bounties) to anyone who detects and responsibly discloses a vulnerability in their products. With the emergence of new AI-specific vulnerabilities, some questions arise:

- Are existing vulnerability bounties likely to extend to AI technologies?
- Should we expect, or encourage, AI vendors to offer bounties for AI-specific exploits?
- Is there scope to offer bounties by third parties (e.g. government, NGO, or philanthropic source) in cases where vendors are unwilling or unable to offer them, for example in the case of popular machine learning frameworks developed as open-source projects or in academia?

## Security Tools

In the same way software development and deployment tools have evolved to include an increasing array of security-related capabilities (testing, fuzzing, anomaly detection, etc.), could we start envisioning tools to test and improve the security of AI components and systems integrated with AI components during development and deployment, such that they are less amenable to attack? These could include:

- Automatic generation of adversarial data
- Tools for analysing classification errors
- Automatic detection of attempts at remote model extraction or remote vulnerability scanning[1]
- Automatic suggestions for improving model robustness (see e.g. Koh and Liang (2017) for related ideas)

1 see e.g. Kesarwani et al., 2017

## Secure Hardware

Hardware innovation has accelerated the pace of innovation in machine learning, by allowing more complex models to be trained, enabling faster execution of existing models, and facilitating more rapid iteration of possible models. In some cases, this hardware is generic (commercial GPUs), but increasingly, AI (and specifically machine learning) systems are trained and run on hardware that is semi-specialized (e.g. graphics processing units (GPUs)) or fully specialized (e.g. Tensor Processing Units (TPUs)). This specialization could make it much more feasible to develop and distribute secure hardware for AI-specific applications than it would be to develop generic secure hardware and cause it to be widely used.

At the workshop we explored the potential value of adding security features to AI-specific hardware. For example, it may be possible to create secure AI hardware that would prevent copying a trained AI model off a chip without the original copy first being deleted. Such a feature could be desirable so that the total number of AI systems (in general or of a certain type or capability level) could be tightly controlled, if the capabilities of such AI systems would be harmful in the wrong hands, or if a large-scale diffusion of such AI systems could have harmful economic, social or political effects.

Other desirable secure hardware features include hardware-level access restrictions and audits. One research trajectory to be considered is developing a reference model for secure AI-specific hardware, which could then be used to inform hardware engineering and, ultimately, be adopted by hardware providers. It may also be the case that potential security threats from AI will drive research in secure hardware more generally, not just for the hardware running AI systems, as a response measure to changes in the cyber threat landscape. Note, however, the potential for manufacturers to undermine the security of the hardware they produce; hardware supply chain vulnerabilities are currently a concern in the cybersecurity context, where there is fear that actors with control over a supply chain may introduce hardware-based vulnerabilities in order to surveil more effectively or sabotage cyber-physical systems[1].

Finally, note that for other security-relevant domains such as cryptography, tamper-proof hardware has been developed[2], with features such as tamper evidence (making it clear that tampering has occurred when it has occurred) and obscurity of layout design (such that it is prohibitively difficult to physically examine the workings of the chip in order to defeat it). Tamper-proof hardware could potentially be valuable so that outsiders are unable to

1 U.S. Defense Science Board, 2017

2 Anderson, 2008

discern the inner workings of an AI system from external emission; so that stolen hardware cannot be used to duplicate an AI; and so that organizations can credibly commit to operating a system in a safe and beneficial way by hard-coding certain software properties in a chip that, if tampered with, would break down. However, secure processors tend to cost significantly more than insecure processors[1] and, to our knowledge, have not specifically been developed for AI purposes.

There are many open questions in this domain:

- What, if any, are the specific security requirements of AI systems, in general and in different domains of application?
- Would changes in the risk landscape (as surveyed above) provide sufficient incentive for a major overhaul of hardware security?
- What set of measures (e.g. reference implementation) would encourage adoption of secure hardware?
- What measures, if any, are available to ensure compliance with hardware safety requirements given the international distribution of vendors and competing incentives such as cost, potential for surveillance and legal implications of auditability?
- How applicable are existing secure processor designs to the protection of AI systems from tampering?
- Could/should AI-specific secure processors be developed?
- How could secure enclaves be implemented in an AI context[2]?
- Can secure processors be made affordable, or could policy mechanisms be devised to incentivize their use even in the face of a cost premium?

1 Anderson, 2008

2 as suggested by Stoica et al., 2017

### Pre-Publication Risk Assessment in Technical Areas of Special Concern

By pre-publication risk assessment we mean analyzing the particular risks (or lack thereof) of a particular capability if it became widely available, and deciding on that basis whether, and to what extent, to publish it. Such norms are already widespread in the computer security community, where e.g. proofs of concept rather than fully working exploits are often published. Indeed, such considerations are sufficiently widespread in computer security that they are highlighted as criteria for submission to prestigious conferences[1].

Openness is not a binary variable: today, many groups will publish the source code of a machine learning algorithm without specifying the hyperparameters to get it to work effectively, or will reveal details of research but not give details on one particular component that could be part of a crucial data ingestion (or transformation) pipeline. On the spectrum from a rough idea, to pseudocode, to a trained model along with source code and tutorials/tips on getting it to work well in practice, there are various possible points, and perhaps there are multiple axes (see Figure 3[2]). Generally speaking, the less one shares, the higher the skill and computational requirements there are for another actor to recreate a given level of capability with what is shared: this reduces the risk of malicious use, but also slows down research and places barriers on legitimate applications.

For an example of a potentially abusable capability where full publication may be deemed too risky, voice synthesis for a given target speaker (as will reportedly soon be available as a service from the company Lyrebird[3]) is ripe for potential criminal applications, like automated spearphishing (see digital security section) and disinformation (see political security section). On the other hand, as is the case with other technologies with significant potential for malicious use, there could be value in openness for security research, for example in white hat penetration testing.

As described in the Rethinking Openness section of the report, there are clear benefits to the level of openness currently prevalent in machine learning as a field. The extent to which restrictions on publication would affect these benefits should be carefully considered. If the number of restricted publications is very small (as in biotechnology, for example), this may not be a significant concern. If, however, restricted publication becomes common, as in the case of vulnerability disclosure in cybersecurity research, then institutions would need to be developed to balance the needs of all affected parties. For example, responsible disclosure mechanisms in cybersecurity allow researchers and affected vendors to negotiate a period of time for a discovered vulnerability to be patched before the vulnerability is published. In addition to the commercial interests of vendors and the security needs of users, such schemes often also protect researchers from legal action by vendors. In the case of AI, one can imagine coordinating institutions that will withhold publication until appropriate safety measures, or means of secure deployment, can be developed, while allowing the researchers to retain priority claims and gain credit for their work. Some AI-related discoveries, as in the case of

1 see e.g. NDSS, 2018

2

3 Lyrebird, 2017

Figure 3. A schematic illustration of the relationship between openness about an AI capability and the skill required to reproduce that capability.

Decreasing skill requirement

Vague description of achievement

Pseudocode

Source code, trained models, tutorials

Increasing openness

adversarial examples in the wild, may be subsumed under existing responsible disclosure mechanisms, as we discuss below in "Responsible AI 0-day Disclosure".

Some valuable questions for future research related to pre-publication research assessment include:

- What sorts of pre-publication research assessment would AI researchers be willing to consider? To what extent would this be seen as conflicting with norms around openness?
- What can be learned from pre-publication risk assessment mechanisms in other scientific/technological domains?
- Is it possible to say, in advance and with high confidence, what sorts of capabilities are ripe for abuse?
- What sort of heuristics may be appropriate for weighing the pros and cons of opening up potentially-abusable capabilities?
- How can such assessment be incorporated into decision-making (e.g. informing one's openness choices, or incorporating such analysis into publications)?
- Can we say anything fine-grained yet generalizable about the levels of skill and computational resources required to recreate capabilities from a given type (code, pseudocode, etc.) of shared information?
- How does the community adopt such a model in the absence of regulation?

### Central Access Licensing Models

Another potential model for openness is the use of what we call central access licensing. In this model, users are able to access certain capabilities in a central location, such as a collection of remotely accessible secure, interlinked data centers, while the underlying code is not shared, and terms and conditions apply to the use of the capabilities. This model, which is increasingly adopted in industry for AI-based services such as sentiment analysis and image recognition, can place limits on the malicious use of the underlying AI technologies. For example, limitations on the speed of use can be imposed, potentially preventing some large-scale harmful applications, and terms and conditions can explicitly prohibit malicious use, allowing clear legal recourse.

Centralised access provides an alternative to publication that allows universal access to a certain capability, while keeping the underlying technological breakthroughs away from bad actors (though also from well-intentioned researchers). Note though that black box model extraction[1] may allow bad actors to gain access to the underlying technology.

Additionally, similarly to early proposals for, in effect, an information processing "tax" on emails in order to disincentivize spam[2], centralized AI infrastructures better enable constraints to be placed on the use of AI services, such that large-scale attacks like automated spear phishing could be made less economical (though see Laurie and Clayton, 2004 for a criticism of this approach, and Liu and Camp, 2006 for further discussion; the increased interest in crypto-economics following the success of bitcoin may lead to advances in this area).

Finally, note that the concentration of AI services in a particular set of organizations may heighten potential for malicious use at those organizations, including by those acting with the blessing of the relevant organization as well as by insider threats. Indeed, some workshop attendees considered these risks from concentration of power to be the biggest threat from AI technologies; note, however, that in this report we have decided to focus on direct malicious use risks, rather than systemic threats (see Scope). In addition to monopolistic behavior, there are more subtle risks such as the introduction of "backdoors" into machine learning systems that users may be unaware of[3].

Some initial research questions that arise related to a central access licensing model:

- What sorts of services might one want only available on a per-use basis?
- How effectively can a service provider determine whether AI uses are malicious?
- How can a user determine whether a service provider is malicious[4]?
- Is the proposal technologically, legally and politically feasible?
- Who might object to a centralised access model and on what grounds?
- Is there enough of a technology gap such that actors without access cannot develop the technologies independently?

1 Bastani et al., 2017

2 Dwork and Naor, 1993

3 Gu et al., 2017

4 see e.g. Ghodsi et al., 2017

- What are potential risks and downsides to centralised access, e.g. in aggravating political security risks?
- How effective can black box model extraction be in different contexts?
- How useful are limits on the amount or frequency of queries to models as a countermeasure against model inversion (extracting the training data from the model; Fredrikson et al., 2015) and other forms of attack such as membership inference (ascertaining whether certain data is contained in the training data[1])?
- What would be the associated trade-offs of such limits?
- (How) can cloud providers vet the safety or security of AI systems without inspecting their internal workings, if such information is private?
- Are cloud computing providers sufficiently flexible in their services to allow the experimentation required by researchers, or would this intervention be most applicable to preventing potentially harmful dissemination trained AI systems?

1 Shokri et al., 2016

### Sharing Regimes that Favor Safety and Security

One possible approach for reducing security risks of AI is to selectively share certain capability information and data with trusted parties. A somewhat analogous approach is used in the cyber domain — Information Sharing and Analysis Centers (ISACs) and Information Sharing and Analysis Organizations (ISAOs) — where companies share information about cyber attacks amongst themselves. Antivirus and large tech companies themselves serve as points of concentration of knowledge sharing, giving them advantages over other kinds of actors. In the case of AI, one might imagine an arrangement where some particularly powerful or hazardous capabilities (e.g. ones that lend themselves straightforwardly to automated hacking) are shared only with organizations or individuals that meet certain criteria, such as having established safety and security routines, or agreeing to random inspection by other members of the group, or a third-party agency that the group has mutually agreed has oversight and inspection powers over them.

Such an approach might be valuable for facilitating collaborative analysis of safety and security issues, and thus getting some fraction of the benefit of an open source approach (where an even larger number of “eyes” are on the problem), while reducing some risks associated with diffusion. If, based on such analysis, it is concluded that there is no harm in further diffusion, then the capabilities would be published.

Several questions arise about the above proposals:

- What have been the benefits and limitations of existing ISACs and ISAOs and are elements of such models useful to AI?
- What sorts of criteria might be applied to an organization or individual in order to ascertain their trustworthiness to deal with particularly sensitive information?
- What types of information might be shared amongst such a group?
- Should there be a limited-sharing stage for all AI developments, or should capabilities be evaluated individually, and if the latter then on what basis?
- What information types should limited sharing apply to: code, research papers, informal notes?
- How can sufficient trust be established between groups such that this kind of coordination is seen as mutually beneficial?
- Are there any particular incentives which can be created that would make this sort of collaboration more likely (for instance, the creation of a shared cluster to test a certain kind of research on)?
- What are potential risks and downsides to this type of sharing regime?

Note that this mechanism has partial overlap with pre-publication risk assessment in technical areas of special concern and central access licensing model.

### Security, Ethics, and Social Impact Education for Future Developers

There has recently been discussion of the role of ethics education in AI[1], in light of ongoing public and private discussion of the potential and pitfalls of AI. Educational efforts might be beneficial in highlighting the risks of malicious applications to AI researchers, and fostering preparedness to make decisions about when

1 Burton et al., 201

technologies should be open, and how they should be designed, in order to mitigate such risks. As yet there is no long-term research on the impacts of such educational efforts on AI researchers' career development and eventual decision-making, suggesting possible areas for research:

- What are the best practices for ethics and policy education for science and engineering in general that are applicable to AI, especially around mitigating security risks?
- How can ethics education be designed so as to most effectively engage with the interests and concerns of AI developers, rather than being seen as merely a box to be ticked off or a burden unrelated to one's practical decision-making, as sometimes occurs in other domains[1]?
- What ought to be included in such a curriculum: ethical methodologies, principles and/or theories?
- How could such a curriculum be iterated over time as the state of AI and security advances?
- Who would be most effective at providing such a curriculum? Should ethics educators from philosophy and other disciplines be brought in or is it better for the community to develop its own internal capacity to teach AI specific ethics?

1 Sunderland et al., 2013

### Ethics Statements and Standards

Another way of acting on ethical concerns could be multi-stakeholder conversation to develop ethical standards for the development and deployment of AI systems, which could be signed on to by companies, research organizations and others deploying AI systems. Two examples of such processes are the IEEE Global Initiative for Ethical Considerations in Artificial Intelligence and Autonomous Systems[2] and the development of the Asilomar AI Principles[3]. Several questions remain open:

2 IEEE Standards Association, 2017

3 Future of Life Institute, 2017

- What institutional frameworks are appropriate for ensuring that statements and standards concerning ethics are fully implemented in order to ensure that they are more than mere technological 'greenwash'? For instance, should community developed standards include statements about reporting and accountability?

- Should companies and research organizations have a statement on ethics, either taken directly from one of these communal standards or developed in house for their particular situation? If so, how can this be encouraged?
- Are standards and statements of this kind the best way to foster industry-wide conversation about the ethics of AI? What are some alternatives?
- What processes are appropriate for revising and updating ethics statements and standards in order to ensure that they remain flexible and can incorporate best practice whilst retaining their sense of permanence and objectivity.

### Norms, Framings and Social Incentives

As noted in previous sections, there are substantial security risks associated with AI, and in some cases one actor could gain from exploiting such risks. At the same time, there are also substantial upsides to progress in AI research and development, and in many cases AI can be used to enhance rather than diminish security. This raises the questions like the following:

- How can the upsides of AI development be framed in such a way as to galvanize focus on mutually beneficial developments and discourage harmful exploitation?
- What are analogous cases from which lessons can be learned, where a technology that could have been used and thought about in a zero-sum manner was governed in a way that benefited all?
- What processes should be allowed to govern the emergence and implementation of a normative culture for beneficial AI in order both to ensure the creation of strong, enforceable and effective norms, and to avoid this normative culture being used to preserve rigid and/or biased norms that hamper diversity and creativity within the sector?
- What role do diverse normative cultures across fields such as AI development, AI safety and risk management play in both allowing for a diverse range of perspectives to inform public debates about AI and ensuring that more people consider themselves to be ‘insiders’ in such debates, and fewer people consider themselves to be ‘outsiders’

## Technologically Guaranteed Privacy

**Several of the threats within the digital security and political security domains (e.g. automated spear phishing, personalised propaganda) rely on attackers gaining access to private information about individuals. In addition to procedural and legal measures to ensure individuals' privacy, there is increasing research on technological tools for guaranteeing user data privacy, which may also be applicable in the context of AI systems. We highlight two technologies as potentially relevant here: differential privacy-guaranteeing algorithms and secure multi-party computation. There remain open questions regarding both technologies:**

- **Can algorithmic privacy be combined with AI technologies, either in general or in specific domains?**
- **What are the trade-offs, if any, for implementing algorithmic privacy, e.g. in terms of performance or in terms of financial viability of services?**
- **What mechanisms (financial, educational, legal or other) could encourage the adoption of algorithmic privacy in AI systems?**
- **What lessons can be learned by efforts at technologically guaranteed privacy (such as Apple's use of differential privacy)?**

**Differential privacy**

**Many machine learning models are currently being developed by companies for commercial use in APIs (see central access licensing above). Without precautions it is possible for individuals to break anonymity in the underlying dataset of a machine learning model that has been deployed for public use via a model inversion attack[1] or membership inference attack[2]. That is, even without access to the training data, an attacker can in some cases query a model in such a way that information from the underlying data set is revealed.**

**Ji et al. (2014) surveyed methods for providing differential privacy in machine learning systems[3], though they do not address differential privacy in neural networks. Such methods have been reported by, for example, Abadi et al. (2016). In general, differentially private machine learning algorithms combine their training data with noise to maintain privacy while minimizing effects on performance. Generally, differentially private algorithms**

1 Fredrikson et al., 2015

2 Shokri et al., 2016

3 a concept first developed in (Dwork, 2006) referring to strong guarantees on the probability of information leakage lose some performance compared to their non-private equivalents, and so privacy may become a concern if the teams developing models are not incentivized to keep their datasets private.

Secure Multi-Party Computation

Secure multi-party computation (MPC) refers to protocols that allow multiple parties to jointly compute functions, while keeping each party's input to the function private[1]. For instance, one simple MPC protocol allows users to jointly compute the outcome of a vote, without sharing their individual votes with one another.

As an important practical application, MPC protocols make it possible to train machine learning systems on sensitive data without significantly compromising its privacy[2]. For example, medical researchers could train a system on confidential patient records by engaging in an MPC protocol with the hospital that possesses them. A technology company could similarly learn from users' data, in some cases, without needing to access this data.

An active open source development effort (OpenMined) is currently aiming to develop a platform to allow users to sell others the right to train machine learning systems on their data using MPC[3]. A number of other frameworks for privacy-preserving machine-learning have also been proposed[4].

In addition, MPC opens up new opportunities for privacy-preserving web applications and cloud computation. For example, one company may develop machine learning models that can make predictions based on health data. If individuals do not want to send this company copies of their personal medical data, they may instead opt to engage in an MPC protocol with the company, and in particular an MPC protocol where only the individual receives the output. At no point in this process does the company gain any knowledge about the individual's medical data; nevertheless, it is still able to provide its service.

MPC could also help to enable privacy-preserving surveillance[5]. To the extent that AI systems play active roles surveillance, for instance by recognizing faces in videos or flagging suspicious individuals on the basis of their web activity, MPC can be used to increase individual privacy. In particular, MPC makes it possible to operate such systems without needing to collect or access the (often sensitive) data that is being used to make the relevant classifications.

At the same time, the use of MPC protocols remains limited by the fact that, in many cases, they can increase overhead associated

1 Yao, 1982

2 Lindell and Pinkas, 2009

3 OpenMined, 2017

4 e.g. Rouhani et al. (2017)

5 Dowlin et al., 2016; Trask, 2017; Garfinkel, forthcoming

with a computation by multiple orders of magnitude. This means that MPC is best-suited for relatively simple computations or for use cases where increased privacy would be especially valuable.

## Monitoring Resources

One type of measure that might help to predict and/or prevent misuse of AI technology would be to monitor inputs to AI systems. Such monitoring regimes are well-established in the context of other potentially dangerous technologies, most notably the monitoring of fissile materials and chemical production facilities for the purpose of implementing nuclear and chemical weapon agreements. An obvious example of an input that might be possible to monitor is computing hardware. While efforts have been made in the past to survey computing resources[1], there is no major ongoing public effort to do so, with the best available information likely withheld due to commercial or state secrecy. One possible benefit to having a public, or semi-public, database of the global distribution of computing resources could be to better understand the likely distribution of offensive and defensive AI/cybersecurity capabilities. Additionally, having such monitoring in place would be valuable if stronger measures were to be employed, e.g. enforceable limitations on how hardware could be used. Questions for further consideration include:

1 Hilbert and Lopez, 2011

- How feasible would it be to monitor global computing resources?
- Are different domains more or less tractable to monitor, or more or less important for AI capabilities, than others (e.g. should video game consoles be considered, in light of their large share in total computing[2] but limited current role in AI)?
- What could be done with such information?
- Are there drawbacks to such an effort (e.g. in encouraging wasteful “racing” to have the most computing power)?
- Would other AI inputs be better suited to monitoring than computing resources?

2 Hilbert and Lopez, 2011

## Exploring Legal and Regulatory Interventions

Much of the discussion above focuses on interventions that can be carried out by researchers and practitioners within the AI development community. However, there is a broader space of possible interventions, including legal ones, that should be considered. We note that ill-considered government interventions could be counterproductive, and that it is important that the implications of any specific policy interventions in this area should be carefully analyzed. A number of questions concerning the proper scope for government intervention in AI security arise; we list some initial examples here:

- Is there a clear chain of responsibility for preventing AI security-related problems?
- Which government departments, marketplace actors or other institutions would ideally have what responsibilities, and what would the interactions with the academic and industry communities be?
- How suitable would existing institutions be at playing this role, and how much will it require the establishment of new institutions founded on novel principles or innovative structures in order to effectively operate in such an evolving and technical field?
- Are relevant actors speaking to each other, and coordinating sufficiently, especially across political, legal, cultural and linguistic barriers?
- Are liability regimes adequate? Do they provide the right incentives for various actors to take competent defensive measures?
- How prepared does e.g. the US government feel, and how much appetite would there be for focused offices/channels designed to increase awareness and expertise?
- Should governments hold developers, corporations, or others liable for the malicious use of AI technologies (or, explicitly make them exempt from such liability[1])? What other approaches might be considered for pricing AI security-related externalities[2]?
- What are the pros and cons of government policies requiring the use of privacy-preserving machine learning systems or defenses against adversarial examples and other forms of malicious use?
- Are data poisoning and adversarial example attacks aimed at disrupting AI systems subject to the same legal penalties as traditional forms of hacking? If not, should they be (and how

1 Calo, 2011; Cooper, 2013

2 see e.g. Farquhar et al., 2017

can legal but related tactics like search engine optimization be dealt with if so)?

- Should international agreements be considered as tools to incentivize collaboration on AI security?
- What should the AI security community's "public policy model" be - that is, how should we aim to affect government policy, what should the scope of that policy be, and how should responsibility be distributed across individuals, organizations, and governments?
- Should there be a requirement for non-human systems operating online or otherwise interacting with humans (for example, over the telephone) to identify themselves as such (a "Blade Runner law"[1]) to increase political security?
- What kind of process can be used when developing policies and laws to govern a dynamically evolving and unpredictable research and development environment?
- How desirable is it that community norms, ethical standards, public policies and laws all say the same thing and how much is to be gained from different levels of governance to respond to different kinds of risk (e.g. near term/long term, technical safety / bad actor and high uncertainty / low uncertainty risks)?

1 Wu, 2017

It seems unlikely that interventions within the AI development community and those within other institutions, including policy and legal institutions, will work well over the long term unless there is some degree of coordination between these groups. Ideally discussions about AI safety and security from within the AI community should be informing legal and policy interventions, and there should also be a willingness amongst legal and policy institutions to devolve some responsibility for AI safety to the AI community, as well as seeking to intervene on its own behalf. Achieving this is likely to require both a high degree of trust between the different groups involved in the governance of AI and a suitable channel to facilitate proactive collaboration in developing norms, ethics education and standards, policies and laws; in contrast, different sectors responding reactively to the different kinds of pressures that they each face at different times seems likely to result in clumsy, ineffective responses from the policy and technical communities alike. These considerations motivated our Recommendations #1 and #2.

Future of Humanity Institute

University of Oxford

Centre for the Study of Existential Risk

University of Cambridge

Center for a New American Security

Electronic Frontier Foundation

OpenAI

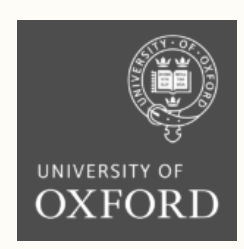

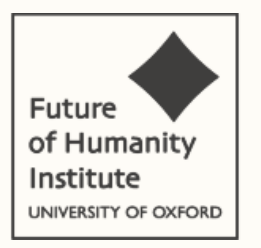

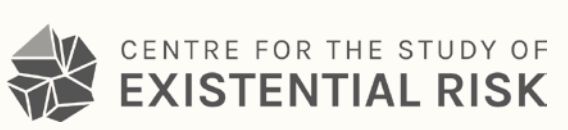

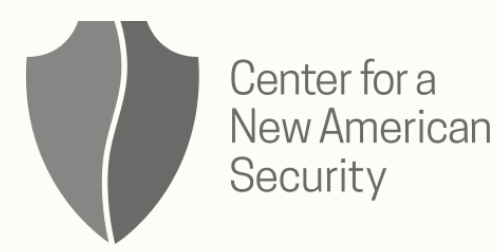

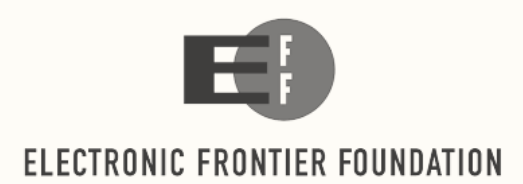

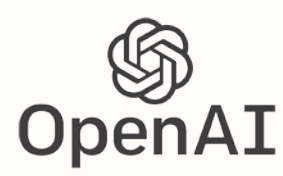

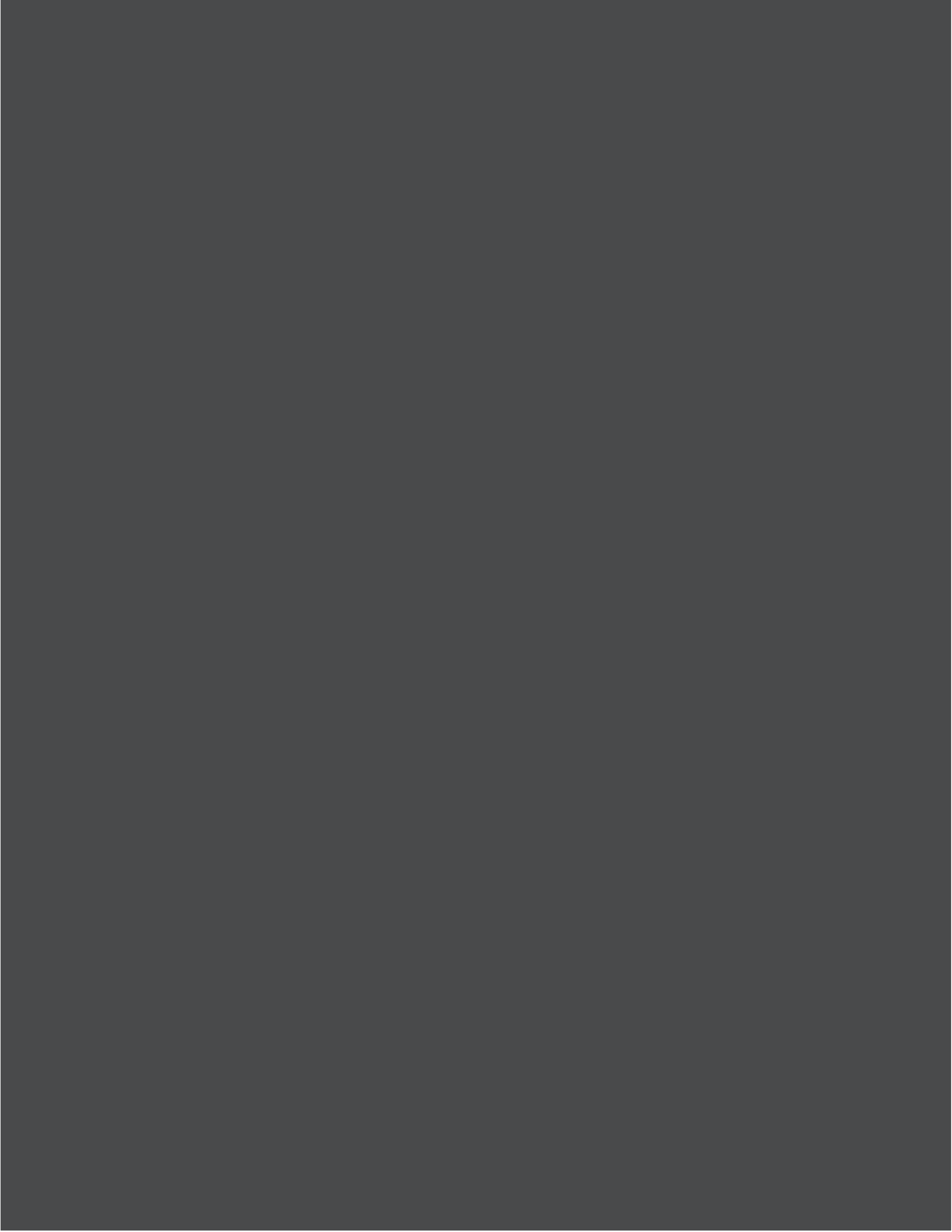